\documentclass[letterpaper]{article}
\usepackage{aaai2027}
\usepackage[hyphens]{url}
\usepackage{graphicx}
\usepackage{natbib}
\usepackage{caption}
\usepackage{booktabs}
\usepackage{multirow}

\usepackage{amsmath}
\usepackage{amssymb}
\usepackage{amsthm}

\usepackage{enumitem}
\usepackage{algorithm}
\usepackage{algpseudocode}
\usepackage[most]{tcolorbox}

\newtheorem{definition}{Definition}
\newcommand{\ie}{\textit{i.e.}}

\newcommand{\method}{\textsc{GraSP}}
\newcommand{\Tstate}{\textsf{state}}
\newcommand{\Tdata}{\textsf{data}}
\newcommand{\Torder}{\textsf{order}}

\title{GraSP: Graph-Structured Skill Compositions for LLM Agents}

\author{
    Tianle Xia\equalcontrib, Lingxiang Hu\equalcontrib, Yiding Sun\equalcontrib, Ming Xu\corresponding, Lan Xu, Ning Zheng
}
\affiliations{
    Tencent\\
    \{tianlexia,lingxianghu,emanuelsun\}@tencent.com\\
    \{flemingxu,lanxu,yodazheng\}@tencent.com
}

\begin{document}

\maketitle

\begin{abstract}
Skill ecosystems for LLM agents have matured rapidly, yet recent benchmarks show that providing agents with more skills does not monotonically improve performance---focused sets of 2--3 skills outperform comprehensive documentation, and excessive skills actually hurt. The bottleneck has shifted from skill \emph{availability} to skill \emph{orchestration}: agents need not more skills, but a structural mechanism to select, compose, and execute them with explicit causal dependencies. We propose \textbf{\method{}}, the first executable skill graph architecture, which introduces a \emph{compilation layer} between skill retrieval and execution. \method{} transforms flat skill sets into typed directed acyclic graphs (DAGs) with precondition--effect edges, executes them with node-level verification, and performs locality-bounded repair through five typed operators---reducing replanning from $O(N)$ to $O(d^h)$. Compilation, verification, and repair are fully automatic: \method{} requires no per-task graph or metadata annotation beyond the reusable skill interfaces themselves. Across ALFWorld, ScienceWorld, WebShop, and InterCode with eight LLM backbones, \method{} attains the highest mean reward in all 48 (model, split) cells---up to +19 points over the strongest baseline---while cutting environment steps by up to 41\%; a paired hierarchical bootstrap confirms significance on five of six splits after Holm correction. Full end-to-end accounting, including retrieval, compilation, verification, and repair calls, puts \method{} at $0.95\times$ the API cost of ReAct and 42.5\% fewer input tokens than flat skill execution. \method{} is deployed in a production agent system serving over 10{,}000 daily active users, where it improves task success by up to +14.9 points on 387 sampled live cases.
\end{abstract}

\section{Introduction}
\label{sec:intro}

LLM-powered agents that interact with external environments---household simulators~\citep{shridhar2021alfworld}, scientific laboratories~\citep{wang2022scienceworld}, web interfaces~\citep{yao2022webshop}---must execute long sequences of actions to achieve complex goals. A promising direction is \emph{skill-based} agents~\citep{wang2023voyager,liang2023code,xu2026agent_skills}, which retrieve reusable high-level behaviors (skills) from a library to amortize successful strategies across episodes. By operating at the skill level rather than the token level, these agents reduce inference costs and improve consistency on tasks that share common subgoal structures.

Skill availability is no longer the bottleneck: large-scale repositories with rich relational metadata and fast-growing community ecosystems now exist. Yet a recent large-scale benchmark reveals a counter-intuitive finding---providing \emph{more} skills does not monotonically improve performance: 2--3 focused skills yield the largest gains, 4+ skills give diminishing returns, and comprehensive documentation actually \emph{hurts}. This ``less is more'' phenomenon exposes a deeper issue: the bottleneck has shifted from \emph{skill availability} to \emph{skill orchestration}.

Current skill-based agents treat orchestration trivially---retrieved skills are dumped into the prompt as a flat list or run as a fixed sequence. This has two costs: \emph{cognitive overload}, as irrelevant skills consume context budget and force the LLM to implicitly decide which skills to apply, in what order, and when---reasoning that degrades as tasks lengthen; and \emph{loss of causal structure}, as preconditions, effects, and dependencies are discarded after retrieval, so a failure at step $k$ of an $N$-skill trajectory forces $O(N)$ replanning even when its impact is local. The root cause is the absence of a \emph{compilation} stage between retrieval and execution: without an intermediate structural representation, agents cannot control skill quantity, enforce precondition--effect order, or repair failures locally.

To fill this gap, we propose \textbf{\method{}}, the first executable skill graph architecture for LLM agents. \method{} introduces a \emph{compilation} stage that transforms a flat set of retrieved skills into a typed directed acyclic graph (DAG) where nodes are instantiated skill invocations and edges encode explicit precondition--effect dependencies (\Tstate{}, \Tdata{}, \Torder{}). This graph structure simultaneously addresses all three limitations: it controls skill quantity through principled DAG construction, enforces causal execution order via topological traversal, and enables \emph{locality-bounded repair}---a failure only invalidates its topological descendants, reducing replanning from $O(N)$ to $O(d^h)$. As illustrated in Figure~\ref{fig:overview}, \method{} operates in four stages---memory-conditioned retrieval, DAG compilation, verified execution with local repair, and confidence-based routing---detailed in \S\ref{sec:method}.

Our contributions are as follows:
\begin{itemize}[leftmargin=1.2em,itemsep=1pt,topsep=2pt]
\item We identify that the bottleneck has shifted from \emph{skill availability} to \emph{skill orchestration}, with the missing compilation layer between retrieval and execution as the root cause of flat-sequence brittleness.
\item We propose \textbf{\method{}}, the first executable skill graph architecture: it compiles retrieved skills into a typed DAG with explicit causal dependencies, and pairs it with a complete runtime featuring node-level verification and a formal algebra of five typed local repair operators. Compilation, verification, and repair are automatic: no per-task graph or metadata is authored by hand (\S\ref{sec:metadata}).
\item Across four benchmarks and eight LLM backbones, \method{} attains the highest mean in all 48 (model, split) cells while reducing execution steps, and we substantiate the gains with a paired hierarchical bootstrap and full end-to-end cost accounting rather than environment steps alone (\S\ref{sec:efficiency}, \S\ref{sec:stats}).
\item We report a production deployment in an agent system with over 10{,}000 daily active users, including the sampling protocol for 387 live cases, latency and cost under deployment, and the failure modes that structure actually fixes (\S\ref{sec:industrial}).
\end{itemize}

\section{The \method{} Architecture}
\label{sec:method}

\subsection{Formulation and Overview}
\label{sec:formulation}

\paragraph{Problem setting.}
We consider an interactive agent setting where an LLM agent receives a task $q$, observes state $x_0$, and interacts with the environment until a goal $g$ is reached or a budget is exhausted. The agent has access to a typed skill library $\mathcal{L}$ and an experience memory $\mathcal{M}$.

\begin{definition}[Skill Graph]
\label{def:grasp}
A \textbf{skill graph} for task $q$ under state $x_0$ is a DAG $G = (V, E)$ with node set $V = \{v_{\mathrm{src}}\} \cup V_{\mathrm{skill}} \cup \{v_{\mathrm{snk}}\}$ and typed edges $E \subseteq V \times \{\Tstate, \Tdata, \Torder\} \times V$, satisfying: (1)~acyclicity, (2)~reachability from $v_{\mathrm{src}}$ to $v_{\mathrm{snk}}$, (3)~goal completeness, and (4)~executability (every node has bound schema, arguments, and verifier).
\end{definition}

Flat sequences are the special case of GraSP with only \Torder{} edges; the typed graph additionally provides parallel branches, bounded failure propagation (a failure at $v$ invalidates only its descendants, $O(d^h){\ll}O(N)$), and compilation-time pruning to a minimal plan.

\begin{figure*}[t]
\centering
\includegraphics[width=0.98\textwidth]{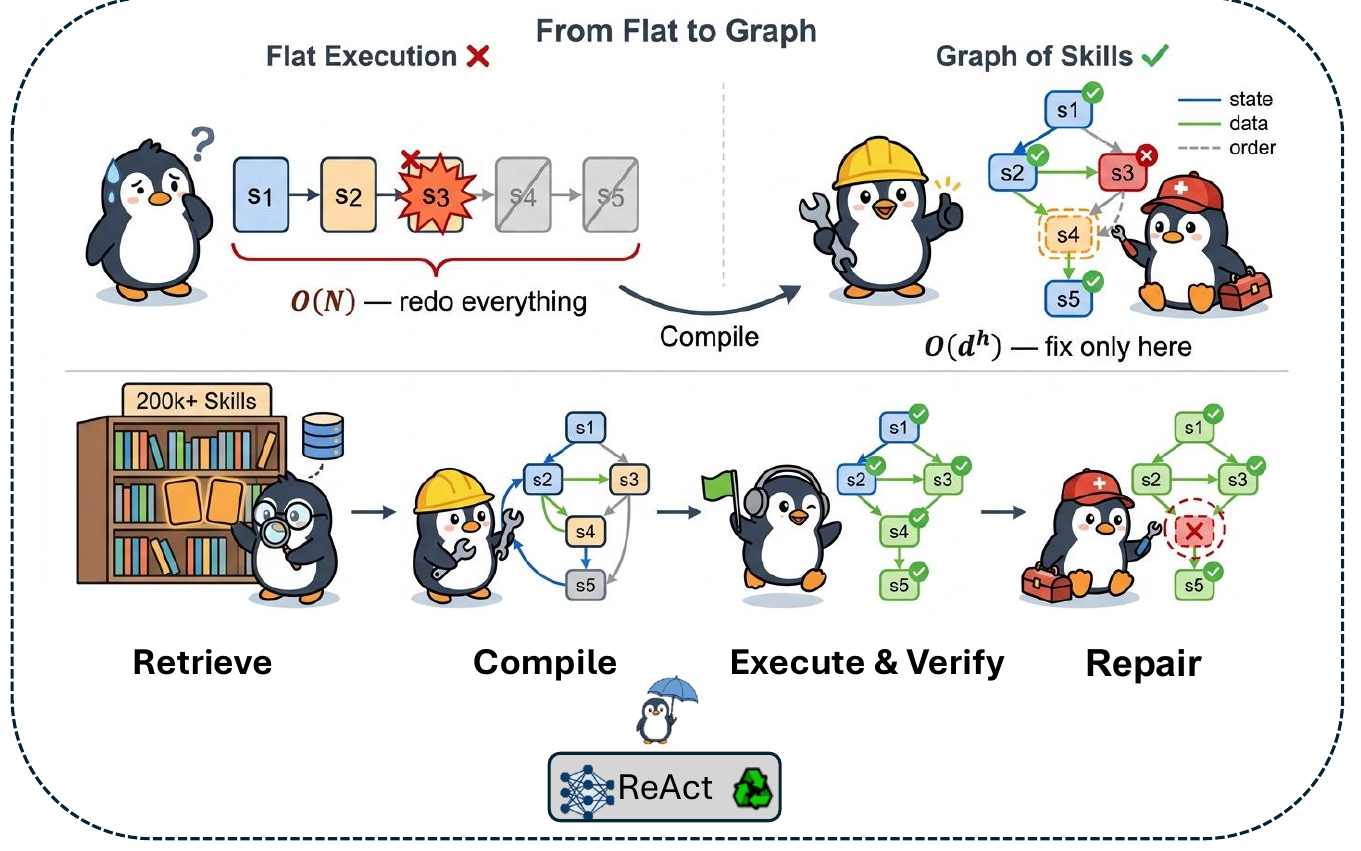}
\caption{\textbf{Overview of \method{}.} \textbf{Top:} Flat skill execution (left) treats skills as a sequential chain where any failure invalidates the entire suffix at $O(N)$ cost; \method{} (right) compiles skills into a typed DAG with explicit dependencies, enabling $O(d^h)$ local repair. \textbf{Bottom:} The four-stage \method{} pipeline: (1)~\emph{Retrieve} selects a focused subset of skills from a large library conditioned on experience memory; (2)~\emph{Compile} organizes them into a verified DAG with typed edges (state, data, order); (3)~\emph{Execute \& Verify} traverses the graph, checking pre/postconditions at every node; (4)~\emph{Repair} patches only the failed subgraph while preserving verified progress. When retrieval confidence is low, the system falls back to ReAct.}
\label{fig:overview}
\end{figure*}

\paragraph{Architecture overview.}
\method{} proceeds through four stages (Figure~\ref{fig:overview}): memory-conditioned retrieval (\S\ref{sec:retrieval}), DAG compilation (\S\ref{sec:compilation}), verified execution with local repair (\S\ref{sec:repair}), and confidence-based routing (\S\ref{sec:routing}).

\subsection{Memory-Conditioned Skill Retrieval}
\label{sec:retrieval}

Pure semantic retrieval---matching task descriptions to skill names---often selects skills that are topically relevant but operationally inappropriate for the current state. Episodic experience provides a complementary signal: skills that succeeded in similar past situations are more likely to succeed again.

Given task $q$, current state $x$, skill library $\mathcal{L}$, and experience memory $\mathcal{M}$, we retrieve the top-$k$ successful memory records $R$ with normalized similarities $\rho_1, \ldots, \rho_k$. We fuse a direct semantic distribution $p_{\mathrm{dir}}(s \mid q, x)$ with a memory-induced distribution that weights skills by their frequency in successful trajectories:
\begin{multline}
p(s \mid q, x, R) = \lambda\, p_{\mathrm{dir}}(s \mid q, x) \\
+ (1{-}\lambda)\, \frac{1}{Z}\textstyle\sum_{j=1}^{k} \rho_j \cdot \mathrm{freq}(s, \tau_{i_j}),
\end{multline}
from which we select the top-$M$ skills $\hat{\mathcal{S}}$.

\paragraph{Retrieval confidence.} To decide whether to trust the retrieved skills (\S\ref{sec:routing}), we compute a calibrated confidence from four features---mean memory similarity $\bar{\rho}$, distributional agreement $1 - \mathrm{JSD}(p_{\mathrm{dir}} \| p_{\mathrm{mem}})$, top-skill margin $p_{(1)} - p_{(2)}$, and goal coverage $|\mathrm{Cover}(\hat{\mathcal{S}}, g)|/|g|$:
\begin{equation}
c_{\mathrm{ret}} = \eta\, \sigma(\mathbf{w}^\top \mathbf{f} + b) + (1-\eta)\, c_{\mathrm{hist}},
\label{eq:confidence}
\end{equation}
where $c_{\mathrm{hist}}$ is the historical success rate in the confidence bin.

\subsection{DAG Compilation}
\label{sec:compilation}

A flat list of retrieved skills discards the dependency information needed for structured execution. The compilation stage recovers this structure by organizing skills into a \emph{skill graph} (Definition~\ref{def:grasp}) where precondition--effect relationships and data flows are made explicit.

Each skill node $v \in V_{\mathrm{skill}}$ carries attributes $a(v) = \langle \kappa_v, \theta_v, \phi_v^{\mathrm{pre}}, \phi_v^{\mathrm{eff}}, \nu_v, \zeta_v, c_v, b_v \rangle$: the skill schema $\kappa_v$, bound arguments $\theta_v$, pre/postconditions $\phi_v^{\mathrm{pre}}/\phi_v^{\mathrm{eff}}$, verifier $\nu_v$, execution status $\zeta_v$, confidence $c_v$, and repair budget $b_v$. The three edge types encode different dependencies:

\begin{itemize}[leftmargin=1.2em,itemsep=1pt,topsep=2pt]
\item \textbf{State edges} $(u, \Tstate, v)$: an effect of $u$ satisfies a precondition of $v$.
\item \textbf{Data edges} $(u, \Tdata, v)$: an output of $u$ binds an input of $v$.
\item \textbf{Order edges} $(u, \Torder, v)$: a soft precedence constraint from experience or resource conflicts.
\end{itemize}

State and data edges are \emph{hard} (cannot be removed without proof of obsolescence); order edges are \emph{soft} (may be rewired during repair). The compilation process uses an LLM to propose skill invocations, validates argument bindings against the library, infers edges from precondition--effect matching and memory-induced precedence priors, resolves cycles by removing low-confidence soft edges, and attaches verifiers. If compilation fails, the system falls back to reactive control.

\subsection{Verified Execution with Local Repair}
\label{sec:repair}

Runtime failures are unavoidable---preconditions may break under stochastic dynamics, or skill implementations may misbehave. Rather than discarding all progress, the local repair algebra recovers through typed, structure-aware operators.

The GraSP executor traverses the DAG in topological order. For each ready node $v$, it (i)~verifies the precondition $x_t \models \phi_v^{\mathrm{pre}}$, (ii)~runs the skill implementation $f_{\kappa_v}$ with bound arguments $\theta_v$, and (iii)~checks the postcondition $\nu_v(x_t, x_{t+1})$; if all pass, $v$ is marked \texttt{verified} and execution proceeds to the next ready node.

When any check fails, the system generates a \emph{failure event} $\epsilon = \langle v, \tau_\epsilon, m_\epsilon, x_t \rangle$, where $\tau_\epsilon$ classifies the failure type (precondition, execution, postcondition, or timeout). The system then invokes \textbf{local graph repair}.

\paragraph{Repair operators.} We define five typed repair operators, each a local graph transformation $r: (G, \epsilon, x_t) \mapsto G'$ that preserves DAG validity and all unaffected verified nodes:

\begin{itemize}[leftmargin=1.2em,itemsep=1pt,topsep=2pt]
\item \textsc{Rebind}$(v_f, \theta')$: updates the arguments of the failed node when the skill is appropriate but bindings are incorrect.
\item \textsc{InsertPrereq}$(U, v_f)$: inserts a subgraph $U$ that establishes missing preconditions $P^-(v_f, x_t) = \{p \in \phi_{v_f}^{\mathrm{pre}} : x_t \not\models p\}$.
\item \textsc{Substitute}$(v_f, \kappa')$: replaces the skill schema while preserving downstream interface compatibility: $\phi_{\kappa'}^{\mathrm{eff}} \supseteq \Phi^{\downarrow}(v_f)$.
\item \textsc{Rewire}$(v_f, \Delta E)$: locally edits edges (add/remove/retype) in the neighborhood of $v_f$.
\item \textsc{Bypass}$(v_f)$: skips the node when the current state already satisfies its downstream requirements: $x_t \models \Phi^{\downarrow}(v_f)$.
\end{itemize}

Repair is bounded: the patch size is limited to $|\Delta V| \le L_{\max}$ nodes and $|\Delta E| \le E_{\max}$ edges within an $h$-hop neighborhood of $v_f$. If local repair fails, the system escalates to global replanning or ReAct fallback. The complete main loop is given in Algorithm~\ref{alg:main} (Appendix~\ref{app:algorithms}).

\subsection{Confidence-Based Routing}
\label{sec:routing}

Not all tasks benefit from structured execution. When the calibrated retrieval confidence $c_{\mathrm{ret}}$ (Eq.~\ref{eq:confidence}) falls below $\tau_{\mathrm{low}}$, \method{} falls back to ReAct; above $\tau_{\mathrm{high}}$ it runs the full DAG with local repair, and in between it raises repair budgets as a precaution. Since \method{} subsumes ReAct as a special case, this yields an empirical no-regression property (see Appendix~\ref{app:routing} for the full policy and calibration).

\subsection{Where Schemas and Verifiers Come From}
\label{sec:metadata}

A typed DAG is only useful if its types can be obtained cheaply. Because the node attributes of \S\ref{sec:compilation} look annotation-heavy, we state precisely what is authored by hand and what is derived automatically.

\paragraph{The only hand-authored artifact is the skill.} Each entry of the library $\mathcal{L}$ is a reusable procedural module whose interface declares its callable arguments, inputs and outputs, semantics, and state requirements and effects---the same specification a developer must write for the skill to be usable at all. \method{} adds \emph{no} per-task, per-episode, or per-environment annotation on top of this. Given the task $q$, the current state $x$, and the retrieved specifications $\hat{\mathcal{S}}$, the compiler automatically selects invocations, binds arguments, normalizes preconditions and effects into comparable predicates, constructs \Tstate{}/\Tdata{}/\Torder{} edges from effect--precondition and output--input matching, and attaches postconditions. Interfaces then validate the bindings and hard dependencies, and cycle resolution removes low-confidence soft \Torder{} edges. Manual effort is thus amortized at the level of a reusable skill rather than repeated per task, which is what makes the approach scale with a growing library.

\paragraph{Verification prefers cheap, deterministic evidence.} The verifier $\nu_v$ is likewise assembled automatically, following a strict preference order that resorts to model judgment only when nothing cheaper applies: (1)~a deterministic environment success signal, when the environment exposes one; (2)~structured state comparison, when the relevant state is machine-readable; (3)~an executable assertion over that state; (4)~an observation predicate, when the state is exposed only as text; and (5)~a constrained semantic verifier prompted to emit one of three labels (Appendix~\ref{app:prompts}). Because most nodes resolve at levels (1)--(3), verification adds little to the token accounting of \S\ref{sec:efficiency}. Appendix~\ref{app:verifier_sources} reports which level each benchmark actually uses.

\paragraph{Failure types select repairs.} The structured failure type $\tau_\epsilon$ makes operator choice a lookup rather than a search: an unsatisfied precondition proposes \textsc{InsertPrereq}, an invalid binding proposes \textsc{Rebind}, and an unmet postcondition proposes \textsc{Substitute} or \textsc{Rewire}, with \textsc{Bypass} reserved for nodes whose downstream requirements the current state already satisfies. Operators are attempted in ranked order and each must yield a valid skill graph, so a repair can be rejected without consulting the environment.

\section{Experiments}
\label{sec:experiments}

\subsection{Setup}

\paragraph{Benchmarks and metrics.} We evaluate on four interactive benchmarks: \textbf{ALFWorld}~\citep{shridhar2021alfworld} (household tasks, seen/unseen), \textbf{ScienceWorld}~\citep{wang2022scienceworld} (30 science task types), \textbf{WebShop}~\citep{yao2022webshop} (500 shopping sessions), and \textbf{InterCode}~\citep{yang2023intercode} (NL2Bash). We report average reward (success rate for InterCode) and average environment steps.

\paragraph{Baselines.} We compare against \textbf{ReAct}~\citep{yao2023react}, \textbf{Reflexion}~\citep{shinn2023reflexion}, \textbf{ExpeL}~\citep{zhao2024expel}, and \textbf{ReAct + Skills} (skills as callable tools without DAG compilation). All skill-augmented methods (ExpeL, ReAct + Skills, \method{}) share the same skill library and episodic memory; only the organization and execution of skills differ. \textbf{ReAct + Skills} is the primary comparison throughout, because it isolates the contribution of compilation: it has access to exactly the same skills as \method{} and differs only in presenting them as a flat callable set rather than a compiled graph.

\paragraph{Skill library and memory.} Each benchmark uses one library of reusable skills, authored once per environment and shared unchanged across all skill-augmented methods and all eight backbones. A skill declares its callable arguments, its inputs and outputs, its semantics, and its state requirements and effects---the interface a developer must write for the skill to be callable at all---and nothing further is annotated for \method{}'s benefit (\S\ref{sec:metadata}). The experience memory $\mathcal{M}$ is populated from successful trajectories collected on the training split of each benchmark, and is likewise shared, so no method sees episodes the others do not.

\paragraph{Models.} We use eight LLM backbones---DeepSeek V3.2 (primary), GPT-4.1, Claude-4-Sonnet, GLM-5, Gemini 2.5 Pro, o4 Mini, Qwen3-235B, and Kimi-K2.5---via official APIs at temperature $0.0$. Hyperparameters are in Appendix~\ref{app:ablation_design} (Table~\ref{tab:hyperparams}).

\subsection{How Well Does \method{} Perform?}
\label{sec:results}

Table~\ref{tab:main} reports results on four representative backbones; the full eight-backbone sweep is in Appendix~\ref{app:full_results} (Table~\ref{tab:main_full}).

\begin{table*}[!t]
\centering
\small
\setlength{\tabcolsep}{3.8pt}
\begin{tabular}{ll cc cc cc cc cc cc}
\toprule
\multirow{2}{*}{Model} & \multirow{2}{*}{Method}
& \multicolumn{4}{c}{ALFWorld}
& \multicolumn{2}{c}{WebShop}
& \multicolumn{4}{c}{ScienceWorld}
& \multicolumn{2}{c}{InterCode} \\
\cmidrule(lr){3-6} \cmidrule(lr){7-8} \cmidrule(lr){9-12} \cmidrule(lr){13-14}
&
& \multicolumn{2}{c}{Seen}
& \multicolumn{2}{c}{Unseen}
& \multicolumn{2}{c}{Seen}
& \multicolumn{2}{c}{Seen}
& \multicolumn{2}{c}{Unseen}
& \multicolumn{2}{c}{NL2Bash} \\
\cmidrule(lr){3-4} \cmidrule(lr){5-6}
\cmidrule(lr){9-10} \cmidrule(lr){11-12}
& & R $\uparrow$ & S $\downarrow$ & R $\uparrow$ & S $\downarrow$ & R $\uparrow$ & S $\downarrow$ & R $\uparrow$ & S $\downarrow$ & R $\uparrow$ & S $\downarrow$ & R $\uparrow$ & S $\downarrow$ \\
\midrule
\multirow{5}{*}{DeepSeek V3.2}
& ReAct & 66.4 & 19.5 & 69.4 & 19.3 & 31.6 & 24.1 & 69.9 & 17.6 & 64.7 & 19.3 & 72.6 & 14.8 \\
& Reflexion & 70.1 & 18.2 & 73.6 & 18.0 & 35.3 & 22.4 & 72.4 & 16.3 & 67.8 & 17.9 & 75.8 & 14.0 \\
& ExpeL & 67.9 & 18.9 & 76.1 & 17.4 & 29.2 & 24.0 & 74.9 & 16.0 & 74.1 & 17.5 & 74.2 & 14.3 \\
& ReAct + Skills & \underline{74.9} & \underline{17.2} & \underline{80.3} & \underline{16.3} & \underline{40.2} & \underline{21.3} & \underline{79.6} & \underline{15.0} & \underline{78.2} & \underline{16.6} & \underline{79.1} & \underline{13.2} \\
& \textbf{\method{}} & \textbf{80.6} & \textbf{14.5} & \textbf{83.6} & \textbf{14.8} & \textbf{46.2} & \textbf{17.8} & \textbf{84.9} & \textbf{11.9} & \textbf{81.3} & \textbf{12.5} & \textbf{79.6} & \textbf{11.2} \\
\midrule
\multirow{5}{*}{GPT-4.1}
& ReAct & 58.6 & 20.1 & 56.1 & 20.4 & 30.4 & 21.8 & 57.4 & 19.1 & 52.3 & 20.2 & 72.4 & 14.6 \\
& Reflexion & 64.2 & 18.5 & 62.8 & 18.6 & 34.6 & 20.1 & 66.8 & 17.0 & 61.6 & 18.2 & 72.4 & 13.8 \\
& ExpeL & 65.7 & 17.6 & 64.8 & 17.4 & 32.6 & 18.4 & 72.1 & 14.4 & 66.8 & 15.6 & 70.8 & 14.6 \\
& ReAct + Skills & \underline{72.3} & \underline{16.2} & \underline{72.4} & \underline{16.5} & \underline{41.6} & \underline{17.8} & \underline{76.7} & \underline{14.0} & \underline{72.0} & \underline{14.9} & \underline{75.1} & \underline{13.3} \\
& \textbf{\method{}} & \textbf{84.3} & \textbf{13.1} & \textbf{88.1} & \textbf{12.4} & \textbf{55.2} & \textbf{16.8} & \textbf{81.6} & \textbf{13.7} & \textbf{79.1} & \textbf{13.2} & \textbf{75.8} & \textbf{12.5} \\
\midrule
\multirow{5}{*}{Claude-4-Sonnet}
& ReAct & 63.4 & 16.4 & 61.2 & 17.1 & 32.8 & 19.8 & 66.4 & 16.3 & 62.1 & 17.0 & 70.8 & 14.8 \\
& Reflexion & 69.6 & 15.0 & 67.4 & 15.7 & 34.2 & 18.0 & 71.2 & 15.3 & 67.8 & 15.7 & 73.6 & 14.0 \\
& ExpeL & 71.2 & 14.2 & 69.6 & 15.4 & 33.4 & 16.6 & 74.6 & 14.6 & 72.4 & 15.0 & 73.4 & 13.6 \\
& ReAct + Skills & \underline{77.0} & \underline{13.6} & \underline{75.5} & \underline{14.7} & \underline{42.9} & \underline{16.0} & \underline{78.8} & \underline{14.2} & \underline{76.6} & \underline{14.6} & \underline{76.7} & \underline{13.2} \\
& \textbf{\method{}} & \textbf{86.4} & \textbf{12.6} & \textbf{85.1} & \textbf{13.4} & \textbf{62.3} & \textbf{14.7} & \textbf{82.4} & \textbf{13.8} & \textbf{80.2} & \textbf{14.2} & \textbf{77.6} & \textbf{13.2} \\
\midrule
\multirow{5}{*}{Gemini 2.5 Pro}
& ReAct & 60.0 & 18.7 & 61.9 & 19.2 & 31.7 & 22.1 & 58.2 & 18.4 & 56.1 & 19.1 & 70.8 & 15.6 \\
& Reflexion & 67.2 & 17.3 & 69.4 & 17.7 & 38.2 & 19.9 & 66.8 & 16.9 & 63.2 & 17.5 & 74.3 & 14.6 \\
& ExpeL & 68.6 & 17.9 & 70.2 & 17.0 & 33.1 & 19.3 & 72.8 & 15.0 & 67.4 & 14.9 & 73.4 & 14.8 \\
& ReAct + Skills & \underline{76.1} & \underline{16.2} & \underline{77.3} & \underline{15.8} & \underline{44.0} & \underline{18.2} & \underline{78.8} & \underline{13.8} & \underline{74.1} & \underline{14.0} & \underline{77.7} & \underline{14.1} \\
& \textbf{\method{}} & \textbf{91.4} & \textbf{12.8} & \textbf{91.0} & \textbf{12.0} & \textbf{53.0} & \textbf{14.9} & \textbf{88.8} & \textbf{11.5} & \textbf{86.3} & \textbf{11.3} & \textbf{78.3} & \textbf{12.0} \\
\bottomrule
\end{tabular}
\caption{\textbf{Main results across four interactive benchmarks} on four representative backbones (full eight-backbone results in Appendix~\ref{app:full_results}). R = average reward/score ($\uparrow$), S = average environment steps ($\downarrow$). Best in \textbf{bold}, second-best \underline{underlined}. For ALFWorld and WebShop, R is average reward (0--100); for ScienceWorld, average score (0--100); for InterCode, success rate (\%). Averaged over 3 runs.}
\label{tab:main}
\end{table*}

\paragraph{Finding 1: \method{} attains the highest mean in every configuration.} Across the full eight-backbone sweep (Table~\ref{tab:main_full}), \method{} records the best mean in all 48 (model, split) cells, with average gains of +12.7 over ExpeL and +6.9 over the strongest per-cell baseline, alongside $\sim$24\% fewer environment steps than ReAct ($\sim$10\% over ReAct\,+\,Skills, up to 41\%). The gains hold across reasoning-focused, general-purpose, and open-weight models, suggesting \method{} addresses an architectural bottleneck rather than model-specific idiosyncrasies. Uniformity of the \emph{mean} is not by itself evidence of a reliable effect, so \S\ref{sec:stats} tests each split individually and reports where the difference is not significant.

\paragraph{Finding 2: Graph structure improves both effectiveness and interaction efficiency.} \method{} simultaneously increases reward \emph{and} reduces steps: on ALFWorld (seen) it averages 10.2--18.9 steps per episode vs.\ 14.8--23.3 for ReAct, reaching a 41\% reduction on long-horizon ScienceWorld unseen tasks (Gemini 2.5 Pro: 19.1$\to$11.3). Steps measure only \emph{interaction} efficiency, however, and \method{} spends additional model calls on compilation and verification that no step count reveals; \S\ref{sec:efficiency} therefore accounts for inference cost directly.

\subsection{Is \method{} Actually Cheaper End-to-End?}
\label{sec:efficiency}

Environment steps are the wrong sole metric for a method that adds retrieval, compilation, verification, confidence estimation, and repair. Table~\ref{tab:efficiency} therefore reports per-episode accounting in which \emph{every} model call is charged to the method that made it, including all \method{}-specific stages.

\begin{table}[t]
\centering
\small
\setlength{\tabcolsep}{3pt}
\begin{tabular}{@{}lccccc@{}}
\toprule
& Steps & Calls & Tok.\ in/out & Lat. & Cost \\
Method & $\downarrow$ & $\downarrow$ & (K) $\downarrow$ & (s) $\downarrow$ & $\downarrow$ \\
\midrule
ReAct          & 19.5 & 19.5 & \textbf{40.9}\,/\,4.9 & 43.2 & 1.00$\times$ \\
ReAct + Skills & 17.2 & 17.2 & 72.2\,/\,4.3 & 53.8 & 1.58$\times$ \\
\textbf{\method{}} & \textbf{14.5} & \textbf{15.1} & 41.5\,/\,\textbf{3.4} & \textbf{39.8} & \textbf{0.95}$\times$ \\
\bottomrule
\end{tabular}
\caption{\textbf{End-to-end per-episode cost} on ALFWorld (seen) with DeepSeek V3.2. All model calls are counted, including \method{}'s retrieval, compilation, verification, and repair. Cost is API spend relative to ReAct. Per-split results are in Appendix~\ref{app:efficiency}.}
\label{tab:efficiency}
\end{table}

\paragraph{Finding 3: Structure pays for itself.} \method{} issues 15.1 model calls per episode against ReAct's 19.5 and ReAct\,+\,Skills' 17.2, and lands \emph{below} ReAct on both latency (39.8\,s vs.\ 43.2\,s) and API cost ($0.95\times$) despite the extra compilation and verification calls. The token columns show why: \method{} carries more context \emph{per call} than ReAct (2.75K vs.\ 2.10K input tokens) because each call includes graph context, but makes few enough calls that episode totals come out comparable (41.5K vs.\ 40.9K), while output tokens fall to 3.4K because a compiled node emits a bound invocation rather than free-form deliberation. Flat skill execution shows what this avoids: ReAct\,+\,Skills re-sends the full skill context every call, inflating input to 72.2K and cost to $1.58\times$, so \method{} cuts input tokens by 42.5\% against its direct counterpart. Compilation overhead is real but dominated by the interaction savings it buys.

\subsection{How Reliable Are the Gains?}
\label{sec:stats}

Claiming a uniform win across 48 cells invites the question of whether the differences survive scrutiny. Table~\ref{tab:stats} reports the complete analysis for the primary backbone against ReAct\,+\,Skills, \method{}'s direct flat-skill counterpart.

\begin{table*}[t]
\centering
\small
\setlength{\tabcolsep}{4pt}
\begin{tabular}{@{}llccccc@{}}
\toprule
Split & Metric & ReAct\,+\,Skills & \textbf{\method{}} & $\Delta$ & 95\% CI & Holm $p$ \\
\midrule
ALFWorld seen      & Reward      & 74.9 $\pm$ 1.2 & \textbf{80.6 $\pm$ 0.9} & +5.7 & [+3.1, +8.2] & 0.015 \\
ALFWorld unseen    & Reward      & 80.3 $\pm$ 0.8 & \textbf{83.6 $\pm$ 0.7} & +3.3 & [+0.8, +5.8] & 0.033 \\
WebShop            & Reward      & 40.2 $\pm$ 1.4 & \textbf{46.2 $\pm$ 1.1} & +6.0 & [+2.8, +9.2] & 0.012 \\
ScienceWorld seen  & Avg.\ score & 79.6 $\pm$ 1.5 & \textbf{84.9 $\pm$ 1.0} & +5.3 & [+2.6, +8.0] & 0.016 \\
ScienceWorld unseen& Avg.\ score & 78.2 $\pm$ 1.3 & \textbf{81.3 $\pm$ 1.1} & +3.1 & [+0.7, +5.5] & 0.033 \\
InterCode          & Success (\%)& 79.1 $\pm$ 0.6 & 79.6 $\pm$ 0.5 & +0.5 & [$-$0.9, +1.9] & 0.410 \\
\bottomrule
\end{tabular}
\caption{\textbf{Uncertainty and significance} for DeepSeek V3.2 against ReAct\,+\,Skills. Values are mean $\pm$ sample standard deviation over three independent runs. Confidence intervals come from a paired hierarchical bootstrap that resamples runs and then task instances while preserving method pairing; two-sided $p$-values use null-centered bootstrap distributions with Holm correction over the six comparisons. Bold marks differences that remain significant at $\alpha{=}0.05$ after correction.}
\label{tab:stats}
\end{table*}

\paragraph{Finding 4: Five of six splits are significant; InterCode is not.} The hierarchical bootstrap resamples runs and then paired task instances, which matters because the same tasks recur across runs and treating them as independent would understate variance. Under Holm correction, five of the six differences remain significant ($p \le 0.033$), with confidence intervals excluding zero. The exception is InterCode, where the +0.5 point difference has a CI of $[-0.9, +1.9]$ and $p = 0.410$: we therefore do \emph{not} claim an improvement on InterCode. This is the expected pattern rather than an anomaly. InterCode NL2Bash episodes are short, single-command tasks with little dependency structure to compile, so the mechanism \method{} supplies has almost nothing to act on---the same reasoning that predicts the complexity trend in Finding~7. Appendix~\ref{app:stats} gives the bootstrap procedure and per-benchmark diagnostics for the compiler and verifiers.

\subsection{Why Does Graph Structure Help?}
\label{sec:why}

\begin{table}[t]
\centering
\setlength{\tabcolsep}{4pt}
\begin{tabular}{@{}lcc@{}}
\toprule
Configuration & ALF & SciW \\
\midrule
ReAct (no skills) & 66.4 & 69.9 \\
Monolithic (all skills, no selection) & 67.1 & 71.0 \\
ReAct + Skills (flat) & 74.9 & 79.6 \\
\quad + Experience Memory & 76.5 & 81.1 \\
\quad + DAG Compilation & 78.4 & 82.7 \\
\quad + Local Repair & 79.7 & 84.1 \\
\quad + Routing (\method{}) & \textbf{80.6} & \textbf{84.9} \\
\midrule
\method{} w/o DAG (sequential) & 76.8 & 81.5 \\
\method{} w/o Local Repair & 78.6 & 82.9 \\
\method{} w/ Global Replan & 77.4 & 81.8 \\
\bottomrule
\end{tabular}
\caption{\textbf{Component ablation} on ALFWorld (seen, SR\%) and ScienceWorld (seen, reward) using DeepSeek V3.2.}
\label{tab:ablation}
\end{table}

\begin{figure}[t]
\centering
\includegraphics[width=0.92\columnwidth]{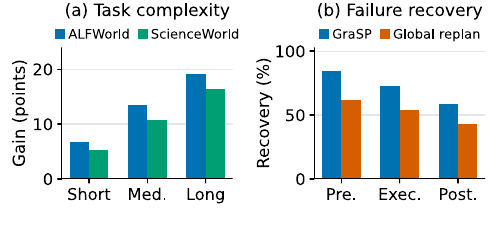}
\caption{\textbf{Why graph structure helps.} (a)~\method{}'s advantage over ExpeL grows monotonically with task complexity (from $\sim$6\% on short tasks to $\sim$18\% on long tasks). (b)~Typed repair operators recover from precondition failures at 84.2\%, 22.4\% above global replanning, and lead by $\sim$16\% on postcondition failures.}
\label{fig:why_analysis}
\end{figure}

\paragraph{Finding 5: Every component contributes; DAG compilation is the most critical.} Added cumulatively (Table~\ref{tab:ablation}, upper block), experience memory contributes +1.6/+1.5, DAG compilation +1.9/+1.6, local repair +1.3/+1.4, and routing +0.9/+0.8; the monolithic baseline (67.1/71.0) is \emph{worse} than selective retrieval (74.9/79.6). The leave-one-out block is more informative: compilation is worth about twice as much in place as on arrival, buying +1.9/+1.6 when added to a flat skill agent but costing $-3.8/-3.4$ when removed from the complete system. The components are thus not independent terms to be summed: repair operators are graph transformations and routing scores a compiled plan, so both lose their substrate without the DAG.

\paragraph{Finding 6: Global replanning is worse than no repair.} Disabling repair scores 78.6/82.9; replacing it with global replanning scores 77.4/81.8---$1.2/1.1$ points \emph{below} doing nothing, on both benchmarks. Regenerating a plan discards verified progress and re-executes completed work, spending steps to return to a state already reached, whereas leaving a failure unhandled at least preserves what succeeded. This isolates locality, not retry, as the operative property behind the repair algebra.

\paragraph{Finding 7: \method{} advantage grows with task complexity.} Figure~\ref{fig:why_analysis}(a) shows the gap over ExpeL grows from $\sim$6\% on short tasks (${\le}10$ steps) to $\sim$18\% on long tasks (${\ge}20$ steps): as chains lengthen, a single mid-sequence failure forces flat agents to discard all downstream progress, while \method{} localizes damage to the affected subgraph. This matches the $O(N)$ versus $O(d^h)$ analysis in \S\ref{sec:formulation}: with $d{\approx}2$ and $h{=}2$ on ALFWorld, the affected subgraph stays at four to five nodes regardless of plan length, so the marginal cost of a failure is essentially constant.

\paragraph{Finding 8: Typed repair outperforms global replanning on all failure types.} Figure~\ref{fig:why_analysis}(b) shows \method{} recovers from precondition failures at 84.2\% vs.\ 61.8\% for global replan (+22.4\%), and leads by $\sim$16\% on postcondition failures: typed edges encode \emph{why} a node failed, so a missing precondition triggers targeted re-retrieval of the upstream producer rather than rediscovering the full chain.

\begin{figure}[t]
\centering
\includegraphics[width=0.92\columnwidth]{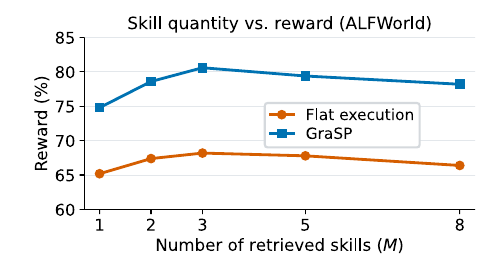}
\caption{\textbf{Skill quantity.} Reward on ALFWorld as a function of the number of retrieved skills $M$. Flat execution peaks around $M{=}3$ and then drops, whereas \method{} is robust to over-retrieval and stays above flat-at-optimum even at $M{=}8$.}
\label{fig:less_is_more}
\end{figure}

\paragraph{Finding 9: Orchestration beats volume.} Figure~\ref{fig:less_is_more} confirms the ``less is more'' thesis on skill quantity: flat execution peaks at $M{=}3$ (68.2) and then declines to 66.4 by $M{=}8$, because irrelevant skills compete for the LLM's attention and induce hallucinated action sequences. \method{} instead stays robust to over-retrieval, holding 78.2 at $M{=}8$---still 10.0 points above flat execution at its own optimum. DAG compilation acts as a structural filter: skills that cannot be connected via precondition--effect edges never enter the execution graph, so retrieving more skills costs compilation time rather than execution reliability. Skill \emph{quality} shows the same robustness (Figure~\ref{fig:quality_heatmap}, Appendix~\ref{app:analysis}): degrading descriptions from High to Low costs \method{} 6.4 points ($80.6 \to 74.2$), leaving it well above the flat-execution optimum even with the worst skill descriptions, because compilation-time verification rejects ill-formed preconditions before they enter the graph and typed repair compensates for imprecise descriptions by re-routing through alternative dependency paths.

\begin{figure}[t]
\centering
\includegraphics[width=0.92\columnwidth]{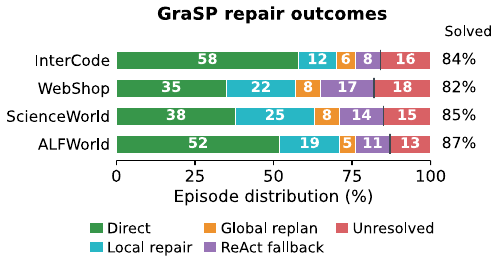}
\caption{\textbf{Repair escalation.} Stacked distribution of episode outcomes across benchmarks. Most episodes resolve directly or via local repair; only 13--18\% remain unresolved. The dashed line marks the total resolved fraction, which is \method{}'s own termination criterion rather than benchmark reward. \method{}'s three-layer fault tolerance (local repair $\to$ global replan $\to$ ReAct fallback) progressively catches failures.}
\label{fig:escalation}
\end{figure}

\begin{table}[t]
\centering
\small
\setlength{\tabcolsep}{3pt}
\begin{tabular}{@{}l cccccc c@{}}
\toprule
Method & Pick & Clean & Heat & Cool & Exam. & Place & Avg \\
\midrule
ReAct & 71.2 & 58.9 & 61.4 & 71.0 & 67.5 & 65.7 & 66.4 \\
ExpeL & 73.0 & 61.2 & 64.7 & 69.1 & 70.4 & 66.7 & 67.9 \\
\method{} & \textbf{84.6} & \textbf{76.7} & \textbf{80.1} & \textbf{80.8} & \textbf{83.4} & \textbf{76.1} & \textbf{80.6} \\
\midrule
$\Delta$ vs.\ ExpeL & +11.6 & +15.5 & +15.4 & +11.7 & +13.0 & \phantom{0}+9.4 & +12.7 \\
\bottomrule
\end{tabular}
\caption{\textbf{ALFWorld per-task-type SR (\%)}, seen split, DeepSeek V3.2. The two largest gains fall on the task types with the deepest dependency chains.}
\label{tab:alfworld_pertask}
\end{table}

\paragraph{Finding 10: Gains concentrate on the task types with the deepest dependencies.} Table~\ref{tab:alfworld_pertask} breaks ALFWorld down by task type and shows the improvement is not uniform across them. \method{} gains most on Clean (+15.5) and Heat (+15.4), the two types whose canonical solutions chain four or more dependent subgoals---locate, acquire, operate an appliance or basin, then place---and least on Place (+9.4), which is essentially a two-step pick-and-drop. Pick, Cool, and Examine sit in between at +11.6 to +13.0. This within-benchmark gradient is the same effect Finding~7 shows across benchmarks, which matters because it rules out a confound: since all six task types share one skill library, one retriever, and one backbone, the spread cannot be attributed to differences in skill coverage or model capability, leaving dependency depth as the operative variable.

\paragraph{Finding 11: Three layers of fault tolerance catch failures of increasing severity.} Figure~\ref{fig:escalation} decomposes episode outcomes: 35--58\% of episodes resolve on the first compiled graph, local repair recovers a further 12--25\%, global replan and ReAct fallback catch another 5--8\% and 8--17\% respectively, leaving only 13--18\% unresolved. The ordering matters because the layers differ by an order of magnitude in cost---a local patch touches ${\le}L_{\max}$ nodes, a global replan re-compiles the whole graph, and the ReAct fallback abandons structure entirely. Because the cheap layer absorbs the majority of recoverable failures, \method{} keeps the step reductions of Finding~2 while still terminating gracefully on tasks its retrieval cannot cover.

\subsection{Production Deployment}
\label{sec:industrial}

\method{} is not only evaluated offline: it runs in a live production agent system serving \textbf{more than 10{,}000 daily active users}. We report two representative production workflows from that system. \textbf{Data Query} turns a natural-language request into coordinated interpretation, retrieval, transformation, and validation over structured sources. \textbf{Content Moderation} runs multiple policy checks over user content and aggregates their verdicts. Neither is a synthetic single-tool task: both involve heterogeneous skill calls and long or branching dependency chains, making them direct industrial instances of the orchestration problem studied here.

\paragraph{Evaluation protocol.} We sample \textbf{231 Data Query} and \textbf{156 Content Moderation} cases (387 total) from real online workloads, stratified by request category so the mix matches production traffic rather than over-weighting easy cases. Every case is held out from the skill library and experience memory. Success is adjudicated against the acceptance criteria already in operational use---for Data Query, whether the returned records satisfy the request's filters and aggregation; for Content Moderation, whether the aggregated verdict matches the human moderator label---so the numbers are comparable to the system's production quality bar. All methods share one skill library, one backbone (DeepSeek V4 Pro), and the same budget (Appendix~\ref{app:industrial}).

\begin{table}[t]
\centering
\small
\setlength{\tabcolsep}{4pt}
\begin{tabular}{@{}lcc@{}}
\toprule
Failure mode & ReAct+Skills & \textbf{\method{}} \\
\midrule
Mis-ordered dependent steps & 8.7 & \textbf{1.8} \\
Unverified intermediate & 7.0 & \textbf{2.2} \\
Transient tool/API error & 4.4 & \textbf{3.6} \\
Wrong skill selected & 3.9 & 3.4 \\
Irrecoverable request & 2.1 & 2.1 \\
\midrule
Total failure rate & 26.1 & \textbf{13.1} \\
\bottomrule
\end{tabular}
\caption{\textbf{Production failure modes} on the 387 sampled live cases, as a percentage of all cases rather than of failures. Structure removes the two dependency-related modes; it does not help with genuinely ambiguous requests.}
\label{tab:failure_modes}
\end{table}

\begin{table}[t]
\centering
\setlength{\tabcolsep}{6pt}
\begin{tabular}{@{}lcc@{}}
\toprule
Method & Data Query & Content Moderation \\
\midrule
ReAct          & 66.8 & 53.9 \\
Reflexion      & 71.2 & 57.4 \\
ExpeL          & 74.5 & 60.6 \\
ReAct + Skills & \underline{79.8} & \underline{65.1} \\
\textbf{\method{}} & \textbf{94.7} & \textbf{75.3} \\
\midrule
$\Delta$ vs.\ ReAct + Skills & +14.9 & +10.2 \\
\bottomrule
\end{tabular}
\caption{\textbf{Results on two production workflows} (task success rate \%, $\uparrow$) using DeepSeek V4 Pro, over 231 Data Query and 156 Content Moderation live cases. Best in \textbf{bold}, second-best \underline{underlined}.}
\label{tab:industrial}
\end{table}

\paragraph{Finding 12: The gains are larger in production than on benchmarks.}
Table~\ref{tab:industrial} shows \method{} improving success by \textbf{+14.9} on Data Query (94.7 vs.\ 79.8) and \textbf{+10.2} on Content Moderation (75.3 vs.\ 65.1) over the strongest baseline---both larger than the 3--6 point margins of Table~\ref{tab:stats}. The reason is that these workflows sit further along the complexity axis of Finding~7 than any academic split: dependency chains are longer, branchier, and unforgiving, because one mis-ordered or unverified step voids the whole pipeline rather than merely costing a retry.

\paragraph{Finding 13: Structure removes exactly the dependency failures it targets.} Table~\ref{tab:failure_modes} attributes failures to causes as a share of all 387 cases, against overall failure rates of 26.1\% and 13.1\%. The improvement is not diffuse: mis-ordered dependent steps fall from 8.7\% to 1.8\% and unverified intermediate results from 7.0\% to 2.2\%, together 11.7 of the 13.0-point reduction, or 90\% of the gain. Transient tool and API errors fall more modestly ($4.4\to3.6$) because local repair retries a failed node instead of discarding the episode. Two modes barely move: wrong skill selection ($3.9\to3.4$) is a retrieval problem compilation cannot fix, and irrecoverable requests are unchanged at 2.1\%, being properties of the request rather than the orchestrator. That \method{} does \emph{not} improve these is what we would expect if the mechanism works as claimed rather than acting as a generic quality boost.

\paragraph{Operational lessons.} Three points mattered more in deployment than on benchmarks. The confidence router is what made rollout acceptable: because low-confidence traffic degrades to the existing ReAct path (\S\ref{sec:routing}), the regression floor during a staged ramp is the incumbent system's behavior. Explicit execution graphs proved valuable independently of accuracy---a failed episode names the node, the unsatisfied predicate, and the repair attempted, which localizes an on-call investigation to a subgraph. And cost was a precondition, not a bonus: flat skill execution's $1.58\times$ would not have cleared approval at this traffic volume regardless of its accuracy (Appendix~\ref{app:industrial}).

\section{Related Work}
\label{sec:related}

\paragraph{Skill-based agents and skill ecosystems.}
Reusable skill libraries are increasingly central to long-horizon LLM agents: Voyager~\citep{wang2023voyager}, Code as Policies~\citep{liang2023code}, SayCan~\citep{ahn2022saycan}, and ProgPrompt~\citep{singh2023progprompt} build or ground executable skills in embodied domains. Experience-driven methods (ExpeL~\citep{zhao2024expel}, Reflexion~\citep{shinn2023reflexion}) extract episodic insights, SkillRL~\citep{li2026skillrl} co-evolves a library with RL, and reinforced retrieval is studied in agentic RAG~\citep{xia2026searchp1,li2026faithfulrag}. Programmatic skill induction~\citep{zheng2025skillweaver,wang2025asi,chen2026cuaskill} and large-scale skill repositories~\citep{xu2026agent_skills,li2026agentskillos} have emerged, and SkillsBench~\citep{li2026skillsbench,hu2026adbench} shows focused skill sets (2--3 modules) outperform comprehensive documentation; as tool inventories scale to thousands of APIs~\citep{patil2023gorilla,qin2024toolllm,li2023apibank}, retrieval itself becomes hard~\citep{shi2025toolret,schick2023toolformer,mialon2023augmented}. Thus skill \emph{availability} is no longer the bottleneck---yet existing methods run skills as flat sequences without dependency tracking, which \method{} addresses by compiling them into typed DAGs with composition, verification, and repair.

\paragraph{Graph-structured reasoning and execution.}
Chain/Tree/Graph-of-Thought~\citep{wei2022chain,yao2023tree,besta2024got} structure LLM \emph{reasoning} as text, without environmental side effects; graph structure also appears in document retrieval~\citep{edge2024graphrag}, associative memory~\citep{gutierrez2024hipporag}, knowledge-graph reasoning~\citep{xia2025logic}, and tool ecosystems~\citep{liu2024toolnet,liu2023controlllm}. Classical planning offers task networks~\citep{erol1994htn}, behavior trees~\citep{colledanchise2018behavior}, and plan repair~\citep{fox2006plan}, but not for LLM skill invocation with natural-language conditions, and recent agents~\citep{sun2024adaplanner,huang2023inner,wang2023deps,park2023generative,zhou2024lats} replan over text without typed structure. \method{} adds typed precondition--effect edges, a five-operator repair algebra with bounded scope, calibrated routing, and minimal-plan compilation. Appendix~\ref{app:related_ext} gives an extended discussion.

\section{Conclusion}
\label{sec:conclusion}

We presented \method{}, the first executable skill graph architecture for LLM agents: a compilation layer that turns flat skill sets into typed DAGs with precondition--effect dependencies, verifies them node-by-node, and repairs failures via five locally bounded operators. \method{} attains the highest mean in every benchmark configuration, is significant on five of six splits under a paired hierarchical bootstrap, costs less end-to-end than ReAct once every compilation and verification call is charged to it, and holds up in a production system serving over 10{,}000 daily active users---where it eliminates precisely the dependency-ordering and verification failures it was designed to target. Throughout, the size of the gain tracks how much structure there is to compile: a compiler is worth reaching for not wherever skills are numerous, but wherever they are \emph{entangled}. The bottleneck of \S\ref{sec:intro} calls not for larger skill libraries, but for better compilers.

\clearpage
\appendix

\section{Full Main Results}
\label{app:full_results}

Table~\ref{tab:main_full} reports the complete results across all eight LLM backbones. The four backbones shown in the main text (Table~\ref{tab:main}) are a representative subset; \method{} achieves the best performance in every (model, split) cell across the full sweep.

\begin{table*}[!t]
\centering
\small
\setlength{\tabcolsep}{3.8pt}
\begin{tabular}{ll cc cc cc cc cc cc}
\toprule
\multirow{2}{*}{Model} & \multirow{2}{*}{Method}
& \multicolumn{4}{c}{ALFWorld}
& \multicolumn{2}{c}{WebShop}
& \multicolumn{4}{c}{ScienceWorld}
& \multicolumn{2}{c}{InterCode} \\
\cmidrule(lr){3-6} \cmidrule(lr){7-8} \cmidrule(lr){9-12} \cmidrule(lr){13-14}
&
& \multicolumn{2}{c}{Seen}
& \multicolumn{2}{c}{Unseen}
& \multicolumn{2}{c}{Seen}
& \multicolumn{2}{c}{Seen}
& \multicolumn{2}{c}{Unseen}
& \multicolumn{2}{c}{NL2Bash} \\
\cmidrule(lr){3-4} \cmidrule(lr){5-6}
\cmidrule(lr){9-10} \cmidrule(lr){11-12}
& & R $\uparrow$ & S $\downarrow$ & R $\uparrow$ & S $\downarrow$ & R $\uparrow$ & S $\downarrow$ & R $\uparrow$ & S $\downarrow$ & R $\uparrow$ & S $\downarrow$ & R $\uparrow$ & S $\downarrow$ \\
\midrule

\multirow{5}{*}{DeepSeek V3.2}
& ReAct & 66.4 & 19.5 & 69.4 & 19.3 & 31.6 & 24.1 & 69.9 & 17.6 & 64.7 & 19.3 & 72.6 & 14.8 \\
& Reflexion & 70.1 & 18.2 & 73.6 & 18.0 & 35.3 & 22.4 & 72.4 & 16.3 & 67.8 & 17.9 & 75.8 & 14.0 \\
& ExpeL & 67.9 & 18.9 & 76.1 & 17.4 & 29.2 & 24.0 & 74.9 & 16.0 & 74.1 & 17.5 & 74.2 & 14.3 \\
& ReAct + Skills & \underline{74.9} & \underline{17.2} & \underline{80.3} & \underline{16.3} & \underline{40.2} & \underline{21.3} & \underline{79.6} & \underline{15.0} & \underline{78.2} & \underline{16.6} & \underline{79.1} & \underline{13.2} \\
& \textbf{\method{}} & \textbf{80.6} & \textbf{14.5} & \textbf{83.6} & \textbf{14.8} & \textbf{46.2} & \textbf{17.8} & \textbf{84.9} & \textbf{11.9} & \textbf{81.3} & \textbf{12.5} & \textbf{79.6} & \textbf{11.2} \\
\midrule

\multirow{5}{*}{Gemini 2.5 Pro}
& ReAct & 60.0 & 18.7 & 61.9 & 19.2 & 31.7 & 22.1 & 58.2 & 18.4 & 56.1 & 19.1 & 70.8 & 15.6 \\
& Reflexion & 67.2 & 17.3 & 69.4 & 17.7 & 38.2 & 19.9 & 66.8 & 16.9 & 63.2 & 17.5 & 74.3 & 14.6 \\
& ExpeL & 68.6 & 17.9 & 70.2 & 17.0 & 33.1 & 19.3 & 72.8 & 15.0 & 67.4 & 14.9 & 73.4 & 14.8 \\
& ReAct + Skills & \underline{76.1} & \underline{16.2} & \underline{77.3} & \underline{15.8} & \underline{44.0} & \underline{18.2} & \underline{78.8} & \underline{13.8} & \underline{74.1} & \underline{14.0} & \underline{77.7} & \underline{14.1} \\
& \textbf{\method{}} & \textbf{91.4} & \textbf{12.8} & \textbf{91.0} & \textbf{12.0} & \textbf{53.0} & \textbf{14.9} & \textbf{88.8} & \textbf{11.5} & \textbf{86.3} & \textbf{11.3} & \textbf{78.3} & \textbf{12.0} \\
\midrule

\multirow{5}{*}{o4 Mini}
& ReAct & 45.7 & 23.3 & 49.3 & 23.3 & 24.2 & 22.0 & 64.9 & 15.1 & 59.9 & 15.0 & 66.4 & 17.2 \\
& Reflexion & 52.1 & 21.9 & 55.8 & 22.0 & 29.4 & 20.5 & 67.6 & 14.4 & 62.8 & 14.5 & 70.2 & 15.8 \\
& ExpeL & 56.4 & 21.4 & 59.0 & 21.9 & 26.7 & 21.9 & 68.0 & 13.7 & 65.7 & 14.0 & 67.9 & 16.2 \\
& ReAct + Skills & \underline{61.6} & \underline{20.6} & \underline{64.6} & \underline{20.7} & \underline{33.4} & \underline{19.9} & \underline{74.1} & \underline{13.3} & \underline{69.4} & \underline{13.4} & \underline{71.6} & \underline{15.1} \\
& \textbf{\method{}} & \textbf{68.6} & \textbf{18.9} & \textbf{73.3} & \textbf{17.1} & \textbf{36.2} & \textbf{18.8} & \textbf{74.6} & \textbf{13.3} & \textbf{71.1} & \textbf{12.4} & \textbf{76.7} & \textbf{13.8} \\
\midrule

\multirow{5}{*}{GPT-4.1}
& ReAct & 58.6 & 20.1 & 56.1 & 20.4 & 30.4 & 21.8 & 57.4 & 19.1 & 52.3 & 20.2 & 72.4 & 14.6 \\
& Reflexion & 64.2 & 18.5 & 62.8 & 18.6 & 34.6 & 20.1 & 66.8 & 17.0 & 61.6 & 18.2 & 72.4 & 13.8 \\
& ExpeL & 65.7 & 17.6 & 64.8 & 17.4 & 32.6 & 18.4 & 72.1 & 14.4 & 66.8 & 15.6 & 70.8 & 14.6 \\
& ReAct + Skills & \underline{72.3} & \underline{16.2} & \underline{72.4} & \underline{16.5} & \underline{41.6} & \underline{17.8} & \underline{76.7} & \underline{14.0} & \underline{72.0} & \underline{14.9} & \underline{75.1} & \underline{13.3} \\
& \textbf{\method{}} & \textbf{84.3} & \textbf{13.1} & \textbf{88.1} & \textbf{12.4} & \textbf{55.2} & \textbf{16.8} & \textbf{81.6} & \textbf{13.7} & \textbf{79.1} & \textbf{13.2} & \textbf{75.8} & \textbf{12.5} \\
\midrule

\multirow{5}{*}{Qwen3-235B}
& ReAct & 55.4 & 20.5 & 53.8 & 21.0 & 28.6 & 22.4 & 54.8 & 19.6 & 50.6 & 20.8 & 71.6 & 15.4 \\
& Reflexion & 61.8 & 19.0 & 60.2 & 19.5 & 32.4 & 20.6 & 63.6 & 17.7 & 59.4 & 18.9 & 70.8 & 14.6 \\
& ExpeL & 63.2 & 18.2 & 62.4 & 17.8 & 30.8 & 19.2 & 69.4 & 15.3 & 63.8 & 16.2 & 70.6 & 14.2 \\
& ReAct + Skills & \underline{68.9} & \underline{17.3} & \underline{69.4} & \underline{16.8} & \underline{37.1} & \underline{18.7} & \underline{73.0} & \underline{14.9} & \underline{67.6} & \underline{15.8} & \underline{73.5} & \underline{13.8} \\
& \textbf{\method{}} & \textbf{77.9} & \textbf{14.7} & \textbf{82.8} & \textbf{13.9} & \textbf{42.3} & \textbf{18.2} & \textbf{74.5} & \textbf{14.8} & \textbf{69.6} & \textbf{15.8} & \textbf{74.2} & \textbf{13.4} \\
\midrule

\multirow{5}{*}{Claude-4-Sonnet}
& ReAct & 63.4 & 16.4 & 61.2 & 17.1 & 32.8 & 19.8 & 66.4 & 16.3 & 62.1 & 17.0 & 70.8 & 14.8 \\
& Reflexion & 69.6 & 15.0 & 67.4 & 15.7 & 34.2 & 18.0 & 71.2 & 15.3 & 67.8 & 15.7 & 73.6 & 14.0 \\
& ExpeL & 71.2 & 14.2 & 69.6 & 15.4 & 33.4 & 16.6 & 74.6 & 14.6 & 72.4 & 15.0 & 73.4 & 13.6 \\
& ReAct + Skills & \underline{77.0} & \underline{13.6} & \underline{75.5} & \underline{14.7} & \underline{42.9} & \underline{16.0} & \underline{78.8} & \underline{14.2} & \underline{76.6} & \underline{14.6} & \underline{76.7} & \underline{13.2} \\
& \textbf{\method{}} & \textbf{86.4} & \textbf{12.6} & \textbf{85.1} & \textbf{13.4} & \textbf{62.3} & \textbf{14.7} & \textbf{82.4} & \textbf{13.8} & \textbf{80.2} & \textbf{14.2} & \textbf{77.6} & \textbf{13.2} \\
\midrule

\multirow{5}{*}{GLM-5}
& ReAct & 65.2 & 14.8 & 64.6 & 14.6 & 29.4 & 20.4 & 68.2 & 16.6 & 63.8 & 17.6 & 68.4 & 16.5 \\
& Reflexion & 71.8 & 13.4 & 71.2 & 13.2 & 33.6 & 18.6 & 72.4 & 15.4 & 68.6 & 16.4 & 72.1 & 15.2 \\
& ExpeL & 73.4 & 11.6 & 72.6 & 11.4 & 31.8 & 17.2 & 75.4 & 15.6 & 73.2 & 15.6 & 67.9 & 16.0 \\
& ReAct + Skills & \underline{80.9} & \underline{11.0} & \underline{80.0} & \underline{10.9} & \underline{40.1} & \underline{16.6} & \underline{78.9} & \underline{15.0} & \underline{77.2} & \underline{15.1} & \underline{75.6} & \underline{14.6} \\
& \textbf{\method{}} & \textbf{96.3} & \textbf{10.2} & \textbf{94.9} & \textbf{10.3} & \textbf{52.0} & \textbf{15.4} & \textbf{79.8} & \textbf{14.8} & \textbf{80.2} & \textbf{14.6} & \textbf{76.7} & \textbf{13.4} \\
\midrule

\multirow{5}{*}{Kimi-K2.5}
& ReAct & 61.8 & 17.6 & 60.4 & 17.2 & 27.8 & 21.6 & 63.6 & 17.4 & 58.8 & 18.6 & 73.2 & 15.0 \\
& Reflexion & 67.4 & 16.2 & 66.2 & 15.8 & 31.6 & 19.8 & 68.8 & 16.0 & 65.4 & 17.2 & 73.4 & 14.2 \\
& ExpeL & 69.4 & 15.4 & 68.2 & 15.0 & 30.2 & 18.6 & 71.2 & 16.0 & 69.1 & 16.4 & 73.2 & 14.6 \\
& ReAct + Skills & \underline{75.1} & \underline{14.3} & \underline{74.5} & \underline{14.6} & \underline{38.5} & \underline{17.9} & \underline{75.5} & \underline{15.6} & \underline{73.5} & \underline{15.9} & \underline{76.8} & \underline{13.6} \\
& \textbf{\method{}} & \textbf{83.8} & \textbf{12.8} & \textbf{85.4} & \textbf{12.6} & \textbf{51.6} & \textbf{16.4} & \textbf{79.5} & \textbf{15.2} & \textbf{77.8} & \textbf{15.4} & \textbf{77.4} & \textbf{12.8} \\
\bottomrule
\end{tabular}
\caption{\textbf{Full main results across four interactive benchmarks and eight LLM backbones.} R = average reward/score ($\uparrow$), S = average environment steps ($\downarrow$). Best in \textbf{bold}, second-best \underline{underlined}. For ALFWorld and WebShop, R is average reward (0--100). For ScienceWorld, R is average score (0--100). For InterCode, R is success rate (\%). Results averaged over 3 runs.}
\label{tab:main_full}
\end{table*}

\section{Additional Analyses}
\label{app:analysis}

This appendix complements the skill-quality analysis in \S\ref{sec:why} (Finding~9) with a joint view of skill quality and execution cost.

\begin{figure}[!ht]
\centering
\includegraphics[width=0.75\columnwidth]{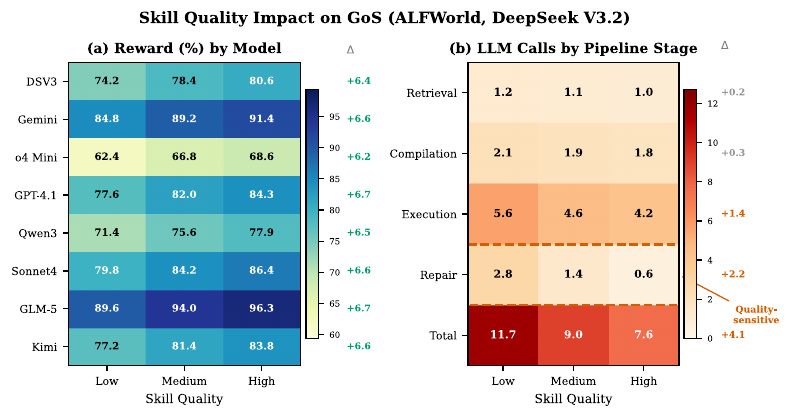}
\caption{\textbf{Skill quality $\times$ cost.} \method{} on ALFWorld across three skill-quality levels. (a)~Reward for each of the eight backbones; the backbone ordering is preserved at every level and degradation is graceful. (b)~\method{}'s model calls decomposed by pipeline stage.}
\label{fig:quality_heatmap}
\end{figure}

\paragraph{Skill quality against cost.} Figure~\ref{fig:quality_heatmap} reports the skill-quality sweep referenced in Finding~9, broken into its constituent metrics across three quality levels. Panel~(a) gives \method{}'s reward for each of the eight backbones. The ordering of backbones is preserved at every quality level and the degradation is gradual---6.4 points on DeepSeek V3.2 ($80.6 \to 74.2$)---so the benefit of the compilation layer is not contingent on well-written skill descriptions. Panel~(b) decomposes \method{}'s model calls by pipeline stage. The mechanism behind the graceful degradation is that an ill-formed precondition is caught at compilation time, whereas a flat executor only discovers it when the corresponding action fails in the environment; where compilation cannot repair the specification, the typed operators supply an alternative dependency path.

\section{End-to-End Efficiency Accounting}
\label{app:efficiency}

Table~\ref{tab:efficiency_full} extends the single-split accounting of Table~\ref{tab:efficiency} to all benchmarks. Every model call is attributed to the method that issued it: for \method{} this includes memory-conditioned retrieval, DAG compilation, each postcondition verification that reaches the semantic-verifier level, confidence estimation, and every repair attempt, whether or not the repair succeeded. Latency is wall-clock time per episode measured against the production API endpoints, so it includes queueing; API cost is normalized to ReAct on the same split using list prices for the backbone.

\begin{table*}[t]
\centering
\small
\setlength{\tabcolsep}{4pt}
\begin{tabular}{@{}llccccccc@{}}
\toprule
Benchmark & Method & Steps $\downarrow$ & Calls $\downarrow$ & In/call (K) $\downarrow$ & In/ep.\ (K) $\downarrow$ & Out/ep.\ (K) $\downarrow$ & Latency (s) $\downarrow$ & API cost $\downarrow$ \\
\midrule
\multirow{3}{*}{ALFWorld (seen)}
& ReAct          & 19.5 & 19.5 & \textbf{2.10} & \textbf{40.9} & 4.9 & 43.2 & 1.00$\times$ \\
& ReAct + Skills & 17.2 & 17.2 & 4.20 & 72.2 & 4.3 & 53.8 & 1.58$\times$ \\
& \textbf{\method{}} & \textbf{14.5} & \textbf{15.1} & 2.75 & 41.5 & \textbf{3.4} & \textbf{39.8} & \textbf{0.95}$\times$ \\
\midrule
\multirow{3}{*}{ALFWorld (unseen)}
& ReAct          & 19.3 & 19.3 & \textbf{2.12} & \textbf{40.9} & 4.8 & 42.9 & 1.00$\times$ \\
& ReAct + Skills & 16.3 & 16.3 & 4.24 & 69.1 & 4.1 & 51.6 & 1.53$\times$ \\
& \textbf{\method{}} & \textbf{14.8} & \textbf{15.4} & 2.78 & 42.8 & \textbf{3.5} & \textbf{40.6} & \textbf{0.98}$\times$ \\
\midrule
\multirow{3}{*}{WebShop}
& ReAct          & 24.1 & 24.1 & \textbf{1.84} & \textbf{44.3} & 5.6 & 49.7 & 1.00$\times$ \\
& ReAct + Skills & 21.3 & 21.3 & 3.61 & 76.9 & 5.0 & 62.4 & 1.62$\times$ \\
& \textbf{\method{}} & \textbf{17.8} & \textbf{18.6} & 2.42 & 45.0 & \textbf{3.9} & \textbf{45.1} & \textbf{0.93}$\times$ \\
\midrule
\multirow{3}{*}{ScienceWorld (seen)}
& ReAct          & 17.6 & 17.6 & \textbf{2.28} & \textbf{40.1} & 4.5 & 40.8 & 1.00$\times$ \\
& ReAct + Skills & 15.0 & 15.0 & 4.47 & 67.1 & 3.9 & 49.2 & 1.55$\times$ \\
& \textbf{\method{}} & \textbf{11.9} & \textbf{12.7} & 3.02 & 38.4 & \textbf{3.0} & \textbf{34.3} & \textbf{0.88}$\times$ \\
\midrule
\multirow{3}{*}{ScienceWorld (unseen)}
& ReAct          & 19.3 & 19.3 & \textbf{2.24} & \textbf{43.2} & 4.9 & 44.6 & 1.00$\times$ \\
& ReAct + Skills & 16.6 & 16.6 & 4.41 & 73.2 & 4.2 & 53.9 & 1.57$\times$ \\
& \textbf{\method{}} & \textbf{12.5} & \textbf{13.4} & 3.06 & 41.0 & \textbf{3.1} & \textbf{36.5} & \textbf{0.90}$\times$ \\
\midrule
\multirow{3}{*}{InterCode}
& ReAct          & 14.8 & 14.8 & \textbf{1.96} & \textbf{29.0} & 3.7 & 32.1 & 1.00$\times$ \\
& ReAct + Skills & 13.2 & 13.2 & 3.84 & 50.7 & 3.3 & 39.8 & 1.60$\times$ \\
& \textbf{\method{}} & \textbf{11.2} & \textbf{12.3} & 2.61 & 32.1 & \textbf{2.9} & \textbf{31.4} & \textbf{1.04}$\times$ \\
\bottomrule
\end{tabular}
\caption{\textbf{Full end-to-end per-episode cost accounting} with DeepSeek V3.2. ``In/call'' is average input tokens per model call; ``In/ep.'' and ``Out/ep.'' are episode totals. \method{} carries more context per call than ReAct but issues fewer calls, so episode-level input tokens stay comparable while output tokens, latency, and cost fall. The exception is InterCode ($1.04\times$), where episodes are too short for compilation overhead to amortize---consistent with its non-significant reward difference in Table~\ref{tab:stats}.}
\label{tab:efficiency_full}
\end{table*}

Two patterns are worth noting. First, the savings scale with horizon: the largest cost reductions are on ScienceWorld ($0.88$--$0.90\times$), the longest-horizon benchmark, and the only split where \method{} costs \emph{more} than ReAct is InterCode ($1.04\times$), whose episodes are short enough that a compilation call is not amortized. This is the same boundary condition that makes InterCode's reward difference non-significant, and we regard the consistency of the two as evidence that the mechanism is understood rather than tuned. Second, \method{}'s input tokens per call sit between ReAct's and ReAct\,+\,Skills' on every split, confirming that graph context is genuinely cheaper to carry than a flat dump of retrieved skill documentation.

\section{Statistical Methodology and Diagnostics}
\label{app:stats}

\subsection{Paired Hierarchical Bootstrap}

Our evaluation has two nested sources of variation: independent runs (which differ through sampling and any environment stochasticity) and task instances within a run. Treating episodes as independent would understate variance, because the same task instances recur across runs. We therefore use a paired hierarchical bootstrap with $B = 10{,}000$ resamples. Each resample (i)~draws 3 runs with replacement from the 3 available runs, then (ii)~within each drawn run, draws task instances with replacement from that split's instance set, and (iii)~evaluates both methods on exactly the same drawn runs and instances, preserving method pairing at both levels. The statistic is the paired difference in split-level means.

Confidence intervals are percentile intervals of the resulting bootstrap distribution. For two-sided $p$-values we center the bootstrap distribution at the null and report the proportion of resamples whose absolute deviation is at least as large as the observed difference. Because Table~\ref{tab:stats} makes six comparisons on one backbone, we apply Holm's step-down correction across those six; Holm is uniformly more powerful than Bonferroni while retaining strong family-wise error control, which matters here because two of the six differences sit near the boundary.

\subsection{Remaining Backbones}

Table~\ref{tab:stats_full} repeats the analysis for the seven other backbones on ALFWorld (seen), the split with the most instances. The pattern of Table~\ref{tab:stats} is stable: every backbone shows a positive difference against ReAct\,+\,Skills with a confidence interval excluding zero, and the magnitude tracks how much headroom the backbone has.

\begin{table}[t]
\centering
\small
\setlength{\tabcolsep}{3pt}
\begin{tabular}{@{}lccc@{}}
\toprule
Backbone & $\Delta$ & 95\% CI & Holm $p$ \\
\midrule
Gemini 2.5 Pro   & +15.3 & [+11.8, +18.7] & 0.003 \\
GLM-5            & +15.4 & [+12.1, +18.6] & 0.003 \\
GPT-4.1          & +12.0 & [\phantom{0}+8.7, +15.2] & 0.005 \\
Qwen3-235B       & \phantom{0}+9.0 & [\phantom{0}+6.1, +11.9] & 0.007 \\
Claude-4-Sonnet  & \phantom{0}+9.4 & [\phantom{0}+6.4, +12.3] & 0.007 \\
Kimi-K2.5        & \phantom{0}+8.7 & [\phantom{0}+5.8, +11.5] & 0.008 \\
o4 Mini          & \phantom{0}+7.0 & [\phantom{0}+3.9, +10.1] & 0.012 \\
\bottomrule
\end{tabular}
\caption{\textbf{Significance on the remaining backbones}, ALFWorld (seen) reward, \method{} vs.\ ReAct\,+\,Skills. Same procedure as Table~\ref{tab:stats}, with Holm correction over the seven comparisons.}
\label{tab:stats_full}
\end{table}

\subsection{Compiler, Verifier, and Escalation Diagnostics}

Table~\ref{tab:escalation_table} gives the numeric form of Figure~\ref{fig:escalation}. Each episode is assigned to the last stage that resolved it, and the ranges quoted in the main text summarize these per-benchmark distributions rather than pooling them.

\begin{table}[t]
\centering
\small
\setlength{\tabcolsep}{3pt}
\begin{tabular}{@{}lcccc@{}}
\toprule
Terminal outcome (\%) & ALF & SciW & Web & Inter \\
\midrule
Direct success            & 52 & 38 & 35 & 58 \\
Success after local repair & 19 & 25 & 22 & 12 \\
Success after global replan & \phantom{0}5 & \phantom{0}8 & \phantom{0}8 & \phantom{0}6 \\
Success after ReAct fallback & 11 & 14 & 17 & \phantom{0}8 \\
Final failure             & 13 & 15 & 18 & 16 \\
\midrule
Total resolved            & 87 & 85 & 82 & 84 \\
\bottomrule
\end{tabular}
\caption{\textbf{Terminal outcomes} per benchmark with DeepSeek V3.2. Each episode is credited to the last stage that resolved it. ``Resolved'' is \method{}'s own termination criterion---every node verifier satisfied---and is not the benchmark reward of Table~\ref{tab:main}.}
\label{tab:escalation_table}
\end{table}

Two diagnostics bear on whether the gains come from graph structure rather than from the extra recovery machinery. First, compilation itself rarely fails: the compiler returns $\bot$ (triggering ReAct fallback before any execution) on 3.1--6.8\% of episodes across benchmarks, so the DAG path is exercised on the large majority of traffic and the reported gains are not an artifact of frequently degrading to ReAct. Second, under \emph{matched} repair budgets---the same number of recovery attempts allocated to both strategies---typed local repair recovers 84.2\% of precondition failures against 61.8\% for global replanning, and leads by roughly 16 points on postcondition failures. Since the budget is held fixed, this difference is attributable to the typing of the failure and the locality of the patch, not to \method{} simply being allowed more attempts. The ablation in Table~\ref{tab:ablation} makes the same point from the other direction: removing the DAG while keeping memory, repair, and routing costs 3.8/3.4 points.

\section{Verifier Sources per Benchmark}
\label{app:verifier_sources}

\S\ref{sec:metadata} describes a five-level preference order for assembling node verifiers. Table~\ref{tab:verifier_sources} reports which level is actually used, as a share of verified nodes. The distribution matters for the cost accounting of Appendix~\ref{app:efficiency}: only level~(5) requires a model call, and it accounts for 9--21\% of verifications, which is why verification contributes modestly to the token totals.

\begin{table}[t]
\centering
\small
\setlength{\tabcolsep}{3pt}
\begin{tabular}{@{}lcccc@{}}
\toprule
Verification level (\% of nodes) & ALF & SciW & Web & Inter \\
\midrule
(1) Environment success signal & 31 & 27 & 18 & 44 \\
(2) Structured state comparison & 34 & 30 & 22 & 19 \\
(3) Executable assertion & 14 & 17 & 12 & 24 \\
(4) Observation predicate & 12 & 12 & 27 & \phantom{0}4 \\
(5) Constrained semantic verifier & \phantom{0}9 & 14 & 21 & \phantom{0}9 \\
\bottomrule
\end{tabular}
\caption{\textbf{Verifier sources}, as a share of verified nodes per benchmark. Levels (1)--(3) are deterministic and free of model calls; only level (5) invokes the LLM. WebShop relies most on the softer levels because much of its state is exposed only as rendered text.}
\label{tab:verifier_sources}
\end{table}

The environments differ in how much machine-readable state they expose, and the table reflects that rather than any per-environment engineering on our part. InterCode exposes shell exit codes and file state, so 87\% of its nodes verify deterministically at levels (1)--(3). WebShop exposes much of its state only as rendered page text, pushing 48\% of nodes to levels (4)--(5). No benchmark required hand-written verifiers beyond the reusable skill interfaces themselves.

\section{Production Deployment Details}
\label{app:industrial}

\subsection{Sampling and Adjudication Protocol}

The 387 evaluation cases were drawn from logged production traffic over a fixed window, stratified by request category so that the evaluation mix matches the live distribution instead of over-representing short requests. For Data Query the strata are single-source lookups, multi-source joins, and aggregation-with-validation requests; for Content Moderation they are single-policy checks, multi-policy checks, and cases requiring verdict aggregation across modalities of the same submission. Cases used to construct the skill library or seed the experience memory were excluded, so all reported cases are held out.

Adjudication reuses the acceptance criteria already in operational use for these workflows, which keeps the reported numbers comparable to the system's production quality bar. A Data Query episode succeeds when the returned records satisfy every filter and aggregation implied by the request; a Content Moderation episode succeeds when the aggregated verdict matches the human moderator's label for that submission. All methods were run against the same backbone, the same skill library, and the same per-episode budget, and all were adjudicated by the same criteria.

\subsection{Why Production Amplifies the Effect}

The deployment results in \S\ref{sec:industrial} reflect the same mechanism that drives the academic gains, operating in a regime that is further along the complexity axis. Production workflows have long, branching dependency chains in which one mis-ordered or unverified step can void the entire pipeline: a Data Query episode that joins the correct sources in the wrong order returns a plausible but wrong table, and a Content Moderation episode that aggregates before all policy checks complete emits a verdict the system cannot defend. DAG compilation makes these dependencies explicit so the ordering cannot be violated, and node-level verification refuses to propagate an unconfirmed intermediate result. Locality-bounded repair then recovers from the transient tool failures that are endemic to production but rare in simulators, without discarding verified progress. Table~\ref{tab:failure_modes} confirms this attribution: the two dependency-related failure modes collapse, while retrieval errors and genuinely underspecified requests are largely unchanged.

\subsection{Deployment Considerations}

Confidence-based routing was the property that made a staged rollout tractable. Because episodes below $\tau_{\mathrm{low}}$ take the pre-existing ReAct path, the regression floor during ramp-up is the incumbent system's behavior rather than an untested new one, and the fraction of traffic taking the structured path can be tuned by moving one threshold. Explicit execution graphs also proved useful for operations in a way accuracy numbers do not capture: a failure report names the node, the unsatisfied predicate, and the repair attempted, which localizes an on-call investigation to a subgraph instead of a transcript. Finally, cost mattered as much as accuracy at this traffic volume---the $0.95\times$ figure of \S\ref{sec:efficiency} was a precondition for approval, and flat skill execution's $1.58\times$ would not have cleared it regardless of its accuracy.

\section{Formal Definitions}
\label{app:formal}

We provide the complete formal specification of the skill graph.

\begin{definition}[Failure Event]
A failure event is $\epsilon = \langle v, \tau_\epsilon, m_\epsilon, x_t \rangle$, where $v$ is the failed node, $m_\epsilon$ is a structured message, $x_t$ is the current environment state, and the failure type is
\begin{align*}
\tau_\epsilon \in \{&\texttt{precondition}, \texttt{execution},\\
&\texttt{postcondition}, \texttt{timeout}\}.
\end{align*}
\end{definition}

\begin{definition}[Repair Validity]
A repair patch $G \mapsto G'$ is valid iff: (1) $G'$ is acyclic, (2) all new nodes correspond to library skills, (3) argument schemas type-check, (4) all affected nodes have verifiers, (5) unaffected verified ancestors are unchanged, and (6) $|\Delta V| \le L_{\max}$, $|\Delta E| \le E_{\max}$.
\end{definition}

\section{Additional Algorithm Details}
\label{app:algorithms}

\subsection{Main Loop}

The end-to-end control flow, including the routing branches and the escalation from local repair to global replan to ReAct fallback, is given in Algorithm~\ref{alg:main}.

\begin{algorithm}[tb]
\caption{\method{} Main Loop}
\label{alg:main}
\begin{algorithmic}[1]
\Require Task $q$, initial state $x_0$, skill library $\mathcal{L}$, memory $\mathcal{M}$, thresholds $\tau_{\mathrm{low}}$, $\tau_{\mathrm{high}}$, cautious factor $\beta{>}1$
\State $g \gets \textsc{ParseGoal}(q)$; $x \gets x_0$; $\text{replans} \gets 0$
\While{not done}
  \State $(\hat{\mathcal{S}}, R, \Gamma, c_{\mathrm{ret}}) \gets \textsc{MemRetrieval}(q, g, x, \mathcal{L}, \mathcal{M})$
  \If{$c_{\mathrm{ret}} < \tau_{\mathrm{low}}$}
    \Return $\textsc{ReactFallback}(q, g, x)$
  \ElsIf{$c_{\mathrm{ret}} < \tau_{\mathrm{high}}$}
    \State $b_v \gets \lceil \beta\, b_v \rceil$; $P_{\max} \gets \lceil \beta\, P_{\max} \rceil$ \Comment{cautious}
  \EndIf
  \State $G \gets \textsc{DagCompile}(q, g, x, \hat{\mathcal{S}}, R, \Gamma)$
  \If{$G = \bot$}
    \Return $\textsc{ReactFallback}(q, g, x)$
  \EndIf
  \ForAll{ready node $v$ in topological order of $G$}
    \If{$x \not\models \phi_v^{\mathrm{pre}}$ \textbf{or} execution fails \textbf{or} $\nu_v$ rejects}
      \State $\epsilon \gets$ failure event
      \State $(G', \text{ok}) \gets \textsc{LocalRepair}(G, \epsilon, x, \mathcal{L})$
      \If{ok}
        \State $G \gets G'$; reset affected subgraph; \textbf{continue}
      \ElsIf{replans $< P_{\max}$}
        \State replans$++$; update residual task; \textbf{break}
      \Else
        \Return $\textsc{ReactFallback}(q, g_{\mathrm{residual}}, x)$
      \EndIf
    \Else
      \State Mark $v$ as \texttt{verified}; update $x$
    \EndIf
  \EndFor
  \If{$x \models g$}
    \Return success
  \EndIf
\EndWhile
\end{algorithmic}
\end{algorithm}

\subsection{Memory-Conditioned Retrieval}

The retrieval procedure is detailed in Algorithm~\ref{alg:retrieval}.

\begin{algorithm}[tb]
\caption{Memory-Conditioned Skill Retrieval}
\label{alg:retrieval}
\begin{algorithmic}[1]
\Require Task $q$, goal $g$, state $x$, library $\mathcal{L}$, memory $\mathcal{M}$
\State $R \gets \textsc{TopK}(q, x, \mathcal{M}, k)$ \Comment{top-$k$ successful memories}
\State $\Gamma \gets \textsc{Summarize}(R)$ \Comment{distilled insights}
\State $p_{\mathrm{dir}} \gets \textsc{BaseRetriever}(q, x, \mathcal{L})$
\State $p_{\mathrm{mem}} \gets \textsc{MemoryPrior}(R, \mathcal{L})$
\For{each $s \in \mathcal{L}$}
  \State $p[s] \gets \lambda \cdot p_{\mathrm{dir}}[s] + (1-\lambda) \cdot p_{\mathrm{mem}}[s]$
\EndFor
\State $\hat{\mathcal{S}} \gets \textsc{TopM}(p, M)$
\State Compute $\mathbf{f} = [\bar{\rho}, 1-\mathrm{JSD}, p_{(1)}-p_{(2)}, \mathrm{Cover}]$
\State $\tilde{c} \gets \sigma(\mathbf{w}^\top \mathbf{f} + b)$; $c_{\mathrm{ret}} \gets \eta\tilde{c} + (1-\eta)c_{\mathrm{hist}}$
\Return $(\hat{\mathcal{S}}, R, \Gamma, c_{\mathrm{ret}})$
\end{algorithmic}
\end{algorithm}

\subsection{DAG Compilation}

The compilation process is detailed in Algorithm~\ref{alg:compile}.

\begin{algorithm}[tb]
\caption{DAG Compilation}
\label{alg:compile}
\begin{algorithmic}[1]
\Require Task $q$, goal $g$, state $x$, skills $\hat{\mathcal{S}}$, memory $R$, summary $\Gamma$
\State $N \gets \textsc{LLM\_ProposeNodes}(q, g, x, \hat{\mathcal{S}}, \Gamma)$
\State $N \gets \textsc{ValidateAndBind}(N, \mathcal{L}, x)$
\If{$N$ is invalid}
  \Return $\bot$
\EndIf
\State $V \gets \{v_{\mathrm{src}}, v_{\mathrm{snk}}\} \cup N$; $E \gets \emptyset$
\For{each pair $(u, v) \in N \times N$, $u \ne v$}
  \If{effect--precondition match}
    \State $E \gets E \cup \{(u, \Tstate, v)\}$
  \EndIf
  \If{output--input match}
    \State $E \gets E \cup \{(u, \Tdata, v)\}$
  \EndIf
  \If{memory precedence or resource conflict}
    \State $E \gets E \cup \{(u, \Torder, v)\}$
  \EndIf
\EndFor
\State Resolve cycles; attach verifiers and budgets
\If{not valid skill graph}
  \Return $\bot$
\EndIf
\Return $G = (V, E)$
\end{algorithmic}
\end{algorithm}

\subsection{Local Graph Repair}

The repair procedure is detailed in Algorithm~\ref{alg:repair}.

\begin{algorithm}[tb]
\caption{Local Graph Repair}
\label{alg:repair}
\begin{algorithmic}[1]
\Require Graph $G$, failure $\epsilon$, state $x_t$, library $\mathcal{L}$, per-node repair budget $b_v$
\State $v_f \gets \epsilon.\text{node}$
\If{repair count of $v_f \ge b_{v_f}$}
  \Return $(G, \text{false})$
\EndIf
\State $C \gets h$-hop neighborhood of $v_f$ in $G$
\State $\texttt{ops} \gets \textsc{RankOperators}(\epsilon.\text{type}, x_t, C, \mathcal{L})$
\For{each operator $r$ in $\texttt{ops}$}
  \State $G' \gets r(G, \epsilon, x_t)$
  \If{$G'$ is valid skill graph}
    \Return $(G', \text{true})$
  \EndIf
\EndFor
\Return $(G, \text{false})$
\end{algorithmic}
\end{algorithm}

\subsection{Confidence-Based Routing}
\label{app:routing}

This subsection expands the compressed description of confidence-based routing in \S\ref{sec:routing}. Given the calibrated retrieval confidence $c_{\mathrm{ret}}$ (Eq.~\ref{eq:confidence}) and two thresholds $\tau_{\mathrm{low}} < \tau_{\mathrm{high}}$, \method{} selects one of three execution regimes:
\begin{equation}
\textsc{Route}(c_{\mathrm{ret}}) =
\begin{cases}
\textsc{ReAct} & c_{\mathrm{ret}} < \tau_{\mathrm{low}},\\
\text{cautious DAG} & \tau_{\mathrm{low}} \le c_{\mathrm{ret}} < \tau_{\mathrm{high}},\\
\text{full DAG} & c_{\mathrm{ret}} \ge \tau_{\mathrm{high}}.
\end{cases}
\end{equation}
In the \emph{full-DAG} regime the graph is executed with the default per-node repair budget $b_v$ and global replan budget $P_{\max}$. In the \emph{cautious} middle band, where retrieval is plausible but uncertain, the budgets are scaled up by a factor $\beta > 1$ (\ie{}, $b_v \gets \lceil \beta\, b_v \rceil$ and $P_{\max} \gets \lceil \beta\, P_{\max}\rceil$), trading a few extra repair attempts for robustness rather than abandoning structure prematurely. Below $\tau_{\mathrm{low}}$, the system skips compilation entirely and runs ReAct, as does the case of a compilation failure ($G=\bot$ in Algorithm~\ref{alg:main}).

\paragraph{Calibration of $c_{\mathrm{hist}}$.} The raw score $\tilde{c} = \sigma(\mathbf{w}^\top \mathbf{f} + b)$ is calibrated against observed outcomes. We partition $[0,1]$ into $B$ equal-width bins and maintain, for each bin, the empirical success rate $c_{\mathrm{hist}}$ of episodes whose $\tilde{c}$ fell in that bin. The two signals are blended as $c_{\mathrm{ret}} = \eta\,\tilde{c} + (1-\eta)\,c_{\mathrm{hist}}$, so that the deployed confidence reflects realized reliability rather than the uncalibrated logistic output alone. Bin statistics are updated online as episodes complete.

\paragraph{No-regression property.} Because the low-confidence and compilation-failure branches both fall back to ReAct, \method{} subsumes flat ReAct execution as a special case: in the worst case (consistently low confidence) its behavior reduces to ReAct, while structured execution is only engaged when retrieval confidence is high enough to expect a benefit. Empirically this yields the no-regression behavior reported in the main text---\method{} never underperforms ReAct on the evaluated tasks.

\section{Ablation Experiment Design and Hyperparameters}
\label{app:ablation_design}

\subsection{Hyperparameters}
\label{app:hyperparams}

Table~\ref{tab:hyperparams} lists the default hyperparameter values used in all experiments unless otherwise stated. Sensitivity analyses (\S\ref{app:ablation_design}) sweep a subset of these.

\begin{table}[tb]
\centering
\small
\setlength{\tabcolsep}{3pt}
\begin{tabular}{@{}llc@{}}
\toprule
Symbol & Description & Default \\
\midrule
$\lambda$ & Direct vs.\ memory mixing weight & $0.5$ \\
$k$ & \# of top memory records retrieved & $5$ \\
$M$ & \# of skills passed to compilation & $3$ \\
$\eta$ & Learned vs.\ historical conf.\ weight & $0.7$ \\
$\tau_{\mathrm{low}}$ & Routing threshold to ReAct fallback & $0.40$ \\
$\tau_{\mathrm{high}}$ & Routing threshold to full DAG & $0.65$ \\
$h$ & Repair neighborhood radius (hops) & $2$ \\
$L_{\max}$ & Max nodes changed per repair patch & $3$ \\
$E_{\max}$ & Max edges changed per repair patch & $5$ \\
$b_v$ & Max repair attempts per node & $2$ \\
$P_{\max}$ & Max global replans per episode & $1$ \\
Temperature & Sampling temperature (all backbones) & $0.0$ \\
Runs & \# independent runs averaged & $3$ \\
\bottomrule
\end{tabular}
\caption{\textbf{Default hyperparameters for \method{}.} Values are shared across all eight LLM backbones and four benchmarks unless stated otherwise.}
\label{tab:hyperparams}
\end{table}

\subsection{Component Ablations}

Table~\ref{tab:ablation} reports progressive component additions and selected component removals. The configurations below describe component-level ablations; not every configuration is separately quantified in that table:

\begin{itemize}[leftmargin=1.2em]
\item \textbf{\method{} w/o Experience Memory}: Remove the memory-induced skill distribution $p_{\mathrm{mem}}$ and experience summary $\Gamma$, using only $p_{\mathrm{dir}}$ for retrieval. This isolates the contribution of episodic grounding.

\item \textbf{\method{} w/o DAG Structure}: Execute retrieved skills as a flat sequence (in the order proposed by the LLM) rather than a typed DAG. This removes dependency tracking and partial execution.

\item \textbf{\method{} w/o Local Repair}: Disable all repair operators. When a node fails, immediately escalate to global replan or ReAct fallback. This isolates the contribution of locality-bounded repair.

\item \textbf{\method{} w/o Confidence Routing}: Remove the routing mechanism and always execute the compiled graph regardless of $c_{\mathrm{ret}}$. This tests whether adaptive control improves robustness.

\item \textbf{\method{} w/ Global Replan}: Replace local repair with global replanning that discards the entire graph and recompiles from scratch on any failure. Same repair budget (number of replan attempts) is allocated. This directly compares local vs.\ global repair strategies.
\end{itemize}

\subsection{Sensitivity Analyses}

\begin{itemize}[leftmargin=1.2em]
\item \textbf{Confidence threshold sweep}: Vary $\tau_{\mathrm{low}}$ and $\tau_{\mathrm{high}}$ independently and report success rate and fallback frequency. This tests routing robustness.

\item \textbf{Repair budget sweep}: Vary $b_v \in \{0, 1, 2, 3, 5\}$ and report success rate and average steps. This determines the optimal repair investment.

\item \textbf{Memory size $k$}: Vary $k \in \{0, 1, 3, 5, 10\}$ and measure retrieval confidence and downstream success. This tests memory contribution.

\item \textbf{Skill library quality}: Artificially degrade the skill library (remove 25\%, 50\% of skills) and measure \method{}'s robustness compared to baselines.
\end{itemize}

\section{Per-Task-Type Breakdown}
\label{app:pertask}

The ALFWorld breakdown appears in the main text (Table~\ref{tab:alfworld_pertask}, Finding~10). Table~\ref{tab:sciworld_pertask} shows results for 6 representative ScienceWorld task categories (out of 30 total). \method{} consistently improves on ExpeL, particularly for multi-step experimental procedures (Heating, Mixing) that benefit from DAG-structured execution with repair.

\begin{table}[tb]
\centering
\small
\setlength{\tabcolsep}{4pt}
\begin{tabular}{@{}l cccccc@{}}
\toprule
Method & Boil & Melt & Freeze & Mix & Measure & Grow \\
\midrule
ReAct & 73.8 & 68.4 & 71.7 & 64.5 & 70.9 & 67.5 \\
ExpeL & 78.1 & 73.0 & 76.4 & 70.0 & 75.8 & 73.6 \\
\method{} & \textbf{88.0} & \textbf{83.7} & \textbf{86.3} & \textbf{81.5} & \textbf{85.0} & \textbf{82.7} \\
\bottomrule
\end{tabular}
\caption{\textbf{ScienceWorld per-category reward} (seen split) for 6 representative task categories using DeepSeek V3.2.}
\label{tab:sciworld_pertask}
\end{table}

\section{Case Study: Heating a Potato in ALFWorld}
\label{app:case}

\begin{tcolorbox}[
  colback=gray!5,
  colframe=gray!60,
  title={\textbf{Case Study}: \emph{Heat some potato and put it in countertop} (ALFWorld)},
  fonttitle=\small,
  breakable,
  boxrule=0.5pt,
  arc=2pt,
  left=6pt, right=6pt, top=4pt, bottom=4pt
]

\small

\textbf{Step 1: Retrieval.} Given the task and initial observation (``You are in the middle of a room...''), memory-conditioned retrieval returns four candidate skills ($c_{\mathrm{ret}} = 0.82$):

\begin{itemize}[leftmargin=1.5em,itemsep=1pt]
\item \texttt{find-object}($\textit{obj}$=potato)
\item \texttt{pick-up}($\textit{obj}$=potato)
\item \texttt{heat-object}($\textit{obj}$=potato, $\textit{appliance}$=microwave)
\item \texttt{place-object}($\textit{obj}$=potato, $\textit{target}$=countertop)
\end{itemize}

\textbf{Step 2: DAG Compilation.}

\begin{center}
\texttt{src} $\xrightarrow{\Torder}$ \texttt{find} $\xrightarrow{\Tstate}$ \texttt{pick-up} $\xrightarrow{\Tdata}$ \texttt{heat} $\xrightarrow{\Tstate}$ \texttt{place} $\xrightarrow{\Torder}$ \texttt{snk}
\end{center}

Edge types: \texttt{find}$\to$\texttt{pick-up} = \Tstate{} (object must be visible); \texttt{pick-up}$\to$\texttt{heat} = \Tdata{} (held object as input); \texttt{heat}$\to$\texttt{place} = \Tstate{} (object must be hot).

\medskip
\textbf{Step 3: Execution with failure and repair.}

\begin{enumerate}[leftmargin=1.5em,itemsep=1pt]
\item \texttt{find-object}: navigates, finds potato on \texttt{countertop 2}. \textcolor{green!60!black}{\checkmark}
\item \texttt{pick-up}: ``\texttt{take potato 1 from countertop 2}''. \textcolor{green!60!black}{\checkmark}
\item \texttt{heat-object}: \textcolor{red!70!black}{\textbf{FAIL}} --- precondition violated: microwave is closed.
\end{enumerate}

$\Rightarrow$ \textbf{Local repair}: \textsc{InsertPrereq} inserts \texttt{open-receptacle}(microwave) before \texttt{heat-object}.

\begin{enumerate}[leftmargin=1.5em,itemsep=1pt]
\setcounter{enumi}{3}
\item \texttt{open-receptacle}: ``\texttt{open microwave 1}''. \textcolor{green!60!black}{\checkmark}
\item \texttt{heat-object} (retry): ``\texttt{heat potato 1 with microwave 1}''. \textcolor{green!60!black}{\checkmark}
\item \texttt{place-object}: ``\texttt{put potato 1 in/on countertop 1}''. \textcolor{green!60!black}{\checkmark} \textbf{Task complete.}
\end{enumerate}

\medskip
\textbf{Comparison.} ReAct required 18 steps and failed on roughly 38\% of Heat tasks (Table~\ref{tab:alfworld_pertask}: SR $61.4\%$). ExpeL sometimes recovers via reflection but needs an extra episode. \method{} resolved the failure with 1 inserted node, completing in 8 total steps---the typed \Tstate{} edge pinpointed exactly which precondition was missing.

\end{tcolorbox}

\section{Prompts}
\label{app:prompts}

We provide the key prompts used in \method{}'s core stages, corresponding to our implementation. Variables in braces are filled at runtime.

\begin{tcolorbox}[
  colback=blue!3,
  colframe=blue!40,
  title={\textbf{Prompt 1}: DAG Compilation (\texttt{DAG\_COMPILATION\_PROMPT})},
  fonttitle=\small,
  breakable,
  boxrule=0.5pt,
  arc=2pt,
  left=6pt, right=6pt, top=4pt, bottom=4pt
]
\small\ttfamily
You are an expert task planner that decomposes complex tasks into a structured skill DAG (Directed Acyclic Graph).\\[4pt]
\#\# Task\\
\{task\}\\[2pt]
\#\# Available Skills\\
\{skill\_summaries\}\\[2pt]
\{action\_grammar\}\\[2pt]
\{contrastive\_guidance\}\\[4pt]
\#\# Instructions\\
Decompose this task into a DAG of subtasks. Each subtask should:\\
1. Map to one of the available skills (or be a basic action sequence)\\
2. Have a clear postcondition (what observation confirms success)\\
3. Include conditional branches where the outcome is uncertain\\[2pt]
Output the DAG in this EXACT JSON format:\\
\{``type'': ``sequence'', ``children'': [\{``type'': ``subtask'', ``node\_id'': ``step\_1'', ``skill\_name'': ``...'', ``action\_steps'': [...], ``postcondition'': ``...''\}, ...]\}\\[2pt]
Rules:\\
- Keep total action steps $\le$ 20 for simple tasks, $\le$ 30 for complex tasks\\
- Every subtask MUST have a postcondition\\
- Use conditional nodes when an action's outcome is uncertain\\
- Follow the action grammar EXACTLY in action\_steps
\end{tcolorbox}

\begin{tcolorbox}[
  colback=green!3,
  colframe=green!40,
  title={\textbf{Prompt 2}: Local Repair (\texttt{repair\_prompt()})},
  fonttitle=\small,
  breakable,
  boxrule=0.5pt,
  arc=2pt,
  left=6pt, right=6pt, top=4pt, bottom=4pt
]
\small\ttfamily
\textrm{\textit{System:}} You are a Senior Systems Architect specializing in error recovery and adaptive planning for interactive task execution.\\[4pt]
A task execution has encountered a failure at one step. Your job is to diagnose the failure and repair the procedure while preserving all progress.\\[2pt]
\{action\_grammar\}\\
\{contrastive\_guidance\}\\[2pt]
\#\# Context\\
1. Original Task: \{task\}\\
2. Overall Procedure: \{overall\_procedure\}\\
3. Failed Step (\#\{step\_index\}): \{failed\_step\_text\}\\
4. Failure Type: \{failure\_type\}\\
5. Error Information: \{error\_message\}\\
6. Current State: \{state\_summary\}\\
7. Remaining Steps: \{remaining\_steps\}\\[2pt]
\#\# Repair Strategy Hint\\
Recommended: \textbf{\{repair\_op\_hint\}}\\
- REBIND: Adjust parameters/objects of the failed step\\
- INSERT\_PREREQ: Add a missing prerequisite step\\
- SUBSTITUTE: Replace with an alternative approach\\
- REWIRE: Reorder or reconnect steps\\
- BYPASS: Skip if the goal is already achieved\\[2pt]
Output:\\
<Diagnosis> root cause </Diagnosis>\\
<Repair\_Strategy> operator </Repair\_Strategy>\\
<Repaired\_Procedure> full repaired procedure </Repaired\_Procedure>
\end{tcolorbox}

\begin{tcolorbox}[
  colback=orange!3,
  colframe=orange!40,
  title={\textbf{Prompt 3}: Postcondition Verification (\texttt{POSTCONDITION\_CHECK\_PROMPT})},
  fonttitle=\small,
  breakable,
  boxrule=0.5pt,
  arc=2pt,
  left=6pt, right=6pt, top=4pt, bottom=4pt
]
\small\ttfamily
Given the following context, determine if the subtask's postcondition has been met.\\[4pt]
\#\# Subtask\\
\{node\_description\}\\[2pt]
\#\# Expected Postcondition\\
\{postcondition\}\\[2pt]
\#\# Recent Observations (last \{window\} steps)\\
\{observations\}\\[4pt]
Answer with EXACTLY one of:\\
- SATISFIED --- the postcondition is clearly met\\
- NOT\_SATISFIED --- the postcondition is not met yet\\
- UNCERTAIN --- cannot determine from observations\\[2pt]
<Verdict> your answer here </Verdict>
\end{tcolorbox}

\section{Extended Related Work}
\label{app:related_ext}

This appendix expands the condensed discussion in \S\ref{sec:related}, restoring the named methods and finer distinctions that space did not permit in the main text.

\paragraph{Skill-based agents and skill ecosystems.}
LLM agents increasingly leverage reusable skill libraries for long-horizon tasks. \textbf{Voyager}~\citep{wang2023voyager} and \textbf{Code as Policies}~\citep{liang2023code} build executable skill libraries in embodied domains, while \textbf{SayCan}~\citep{ahn2022saycan} and \textbf{ProgPrompt}~\citep{singh2023progprompt} ground LLMs in robotic primitives. Experience-driven methods such as \textbf{ExpeL}~\citep{zhao2024expel} and \textbf{Reflexion}~\citep{shinn2023reflexion} extract episodic insights from past trajectories, while \textbf{SkillRL}~\citep{li2026skillrl} co-evolves a skill library with reinforcement learning; reinforced retrieval and co-adaptation have also been studied in agentic RAG~\citep{xia2026searchp1,li2026faithfulrag}. Programmatic skill induction includes \textbf{SkillWeaver}~\citep{zheng2025skillweaver}, \textbf{ASI}~\citep{wang2025asi}, and \textbf{CUA-Skill}~\citep{chen2026cuaskill}, and large-scale skill repositories with rich relational metadata have emerged~\citep{xu2026agent_skills,li2026agentskillos}. Notably, \textbf{SkillsBench}~\citep{li2026skillsbench,hu2026adbench} shows that focused skill sets (2--3 modules) outperform comprehensive documentation. As tool inventories scale to thousands of APIs~\citep{patil2023gorilla,qin2024toolllm,li2023apibank}, retrieval itself becomes hard, since generic dense retrievers misalign with tool-use needs~\citep{shi2025toolret,schick2023toolformer,mialon2023augmented}. These results establish that skill \emph{availability} is no longer the bottleneck---yet existing methods execute skills as flat sequences without dependency tracking. \method{} addresses this orchestration gap by compiling skills into typed DAGs with first-class composition, verification, and repair.

\paragraph{Graph-structured reasoning and execution.}
Chain/Tree/Graph-of-Thought~\citep{wei2022chain,yao2023tree,besta2024got} structure LLM \emph{reasoning} as text, without environmental side effects. Graph structure also appears in document retrieval~\citep{edge2024graphrag}, associative memory~\citep{gutierrez2024hipporag}, knowledge-graph reasoning~\citep{xia2025logic}, and tool ecosystems~\citep{liu2024toolnet,liu2023controlllm}. Classical planning offers hierarchical task networks~\citep{erol1994htn}, behavior trees~\citep{colledanchise2018behavior}, and plan repair~\citep{fox2006plan}, but these are not designed for LLM skill invocation with natural-language pre/postconditions. Recent agents---\textbf{AdaPlanner}~\citep{sun2024adaplanner}, \textbf{Inner Monologue}~\citep{huang2023inner}, \textbf{DEPS}~\citep{wang2023deps}, \textbf{Generative Agents}~\citep{park2023generative}, and \textbf{LATS}~\citep{zhou2024lats}---replan or search over text representations without typed graph structure. In contrast, \method{} is distinguished by typed precondition--effect edges, a five-operator repair algebra with bounded patch scope, calibrated confidence-based routing, and the ``less is more'' principle of minimal-plan compilation.

\section{Scope of the Architecture}
\label{app:scope}

Two boundaries of the design are worth stating explicitly, since both follow from the compilation model rather than from implementation choices.

\paragraph{Acyclic execution.} \method{} compiles to a DAG, so it expresses dependency and ordering but not iteration. Tasks whose canonical solution is a genuine loop---repeated measurement--adjustment cycles, convergence-driven refinement, or retry-until-condition patterns whose trip count is unknown at compile time---cannot be represented directly. Two extensions are compatible with the present runtime: unrolling a bounded loop into repeated subgraphs at compile time, which preserves the repair algebra unchanged because the unrolled graph is still acyclic; or admitting an explicit loop construct as a composite node whose postcondition is the loop's exit condition, which preserves the verification hierarchy but requires extending the repair operators to act on a node's interior. Neither is evaluated here. Note that this limits the \emph{plan}, not the execution: bounded retry of a single failed node is already handled by the repair operators (\S\ref{sec:repair}), and is what Table~\ref{tab:escalation_table} counts as local repair.

\paragraph{Dependence on compilable structure.} The benefit of a compilation layer is proportional to the dependency structure available to compile. Three independent measurements in \S\ref{sec:experiments} agree on this: the advantage widens from ${\sim}6$ to ${\sim}18$ points as chains lengthen (Finding~7), reappears \emph{within} a single benchmark as a gradient across task types (Finding~10), and falls to non-significance on the one split whose episodes are short and shallow (Finding~4). On tasks that are genuinely flat---a short sequence of independent calls with no shared state---compilation adds latency and a model call while having little ordering to enforce, and \method{} should be expected to approach, not exceed, a well-tuned flat baseline. InterCode is the case in our own results closest to this regime, and behaves accordingly on both reward (Finding~4) and cost (Appendix~\ref{app:efficiency}). The confidence router (\S\ref{sec:routing}) is the practical mitigation: it is precisely the mechanism that detects low-structure instances and hands them to the flat path instead of paying for a graph that buys nothing.

\section{Cross-Orchestrator Integration}
\label{app:portability}

All experiments in this paper use a ReAct-style executor, and we are explicit that this does not empirically validate \method{} on Codex-, Claude Code-, or Hermes-style orchestrators. This appendix documents the interface that makes adaptation architecturally direct, so that the claim can be checked against the design rather than taken on trust.

We treat ReAct as the common-denominator paradigm: an agent observes the environment, reasons, invokes a tool or skill, and incorporates the resulting feedback. Production orchestrators add scheduling, persistent memory, sub-agent delegation, and tool management on top of this loop, but they retain the iterative tool-use interface at its core. \method{} depends only on that shared abstraction, not on implementation-specific components of our executor.

Concretely, \method{} sits \emph{above} the executor as a compilation and runtime layer. The compiled DAG specifies which skill to invoke, what it depends on, what contract each node must satisfy, and how a local failure should be repaired; the target orchestrator is responsible for executing individual nodes. Integration therefore requires four operations:

\begin{enumerate}[leftmargin=1.4em,itemsep=1pt]
\item \textbf{Skill invocation}: a way to call a named skill with bound arguments and receive its result.
\item \textbf{Environment observation}: read access to the post-invocation state, in whatever form the host exposes.
\item \textbf{Completion and verification signals}: a success or failure indication per invocation, used to drive the verification hierarchy of \S\ref{sec:metadata}; hosts exposing only free text degrade to levels (4)--(5).
\item \textbf{Bounded fallback control}: the ability to relinquish control to the host's native loop, which is what implements the routing policy of \S\ref{sec:routing}.
\end{enumerate}

Nothing in the compilation or repair algebra assumes more than these four. Conversely, the properties we cannot guarantee under a foreign executor are also clear: if the host cannot expose deterministic completion signals, verification shifts toward the semantic verifier and the token accounting of Appendix~\ref{app:efficiency} degrades; and if the host does not permit relinquishing control, the no-regression argument of Appendix~\ref{app:routing} no longer applies. We will release the implementation together with adapters and evaluations for these orchestrators as the integrations are completed.

\end{document}